\documentclass[journal]{IEEEtran}
\usepackage{tikz}
\usepackage{pgfplots}
\usepackage{amsmath, amssymb, amsfonts, amsthm, mathtools, dsfont}
\usepackage{cite}
\usepackage{textcomp}
\usepackage[dvipsnames]{xcolor}
\usepackage[utf8]{inputenc}
\usepackage{psfrag}
\usepackage{lipsum}
\usepackage{cuted}
\usepackage{stfloats}
\usepackage{siunitx}
\usepackage[ruled,lined,linesnumbered]{algorithm2e}
\usepackage{balance}
\usepackage{scalerel}
\usetikzlibrary{svg.path}
\usepackage{booktabs}
\usepackage{wasysym}
\usepackage[scr=boondox]{mathalpha}
\usepackage{graphicx}
\usepackage{bm}
\usepackage{url}
\usepackage{upgreek}
\usepackage{yfonts}
\usepackage{arydshln}
\usepackage{soul}
\usepackage{multirow}
\usepackage{hyperref}
\usepackage{cleveref}
\usepackage{microtype} 
\allowdisplaybreaks 
\pgfplotsset{compat=newest}
\usepackage{enumitem}
\usepackage{subcaption}

\usetikzlibrary{arrows.meta, positioning, calc, fit, backgrounds}

\newcommand{\ebf}{\mathbf{e}}

\newcommand{\pbf}{\mathbf{p}}
\newcommand{\qbf}{\mathbf{q}}

\newcommand{\ubf}{\mathbf{u}}
\newcommand{\vbf}{\mathbf{v}}
\newcommand{\wbf}{\mathbf{w}}
\newcommand{\xbf}{\mathbf{x}}

\newcommand{\zbf}{\mathbf{z}}

\newcommand{\Ibf}{\mathbf{I}}

\newcommand{\Rbf}{\mathbf{R}}
\newcommand{\Sbf}{\mathbf{S}}

\newcommand{\Bcal}{{\mathcal{B}}}
\newcommand{\Ccal}{{\mathcal{C}}}

\newcommand{\Ecal}{{\mathcal{E}}}

\newcommand{\Jcal}{{\mathcal{J}}}

\newcommand{\Ncal}{{\mathcal{N}}}

\newcommand{\Scal}{{\mathcal{S}}}

\newcommand{\Ucal}{{\mathcal{U}}}

\newcommand{\Ebb}{\mathbb{E}}
\newcommand{\Rbb}{\mathbb{R}}

\newcommand{\Pbb}{\mathbb{P}}

\newcommand{\tr}{{\rm tr}}
\newcommand{\dB}{{\rm dB}}

\newcommand{\norm}[1]{\left\lVert#1\right\rVert}

\DeclareMathAlphabet\mathbfcal{OMS}{cmsy}{b}{n}

\newtheoremstyle{ehsan}
  {}
  {}
  {}
  {\parindent}
  {\itshape}
  {:}
  { }
  {\thmname{#1}\thmnumber{ #2}\textit{\thmnote{ (#3)}}}

\theoremstyle{ehsan}
\newtheorem{theorem}{Theorem}
\newtheorem{remark}{Remark}
\newtheorem{corollary}{Corollary}
\newtheorem{lemma}{Lemma}
\newtheorem{proposition}{Proposition}
\newtheorem*{proofe}{Proof}

\begin{document}

\title{Communication-Efficient Byzantine-Robust Federated Conformal Prediction via Partial Sharing}

\author{Ehsan Lari, Reza Arablouei, and Stefan Werner, \IEEEmembership{Fellow, IEEE}
\thanks{
Manuscript received 20 February 2026; revised 8 July 2026 and 29 August 2026; accepted 20 September 2026.
This work was partially supported by the Research Council of Norway and the Research Council of Finland (Grant 354523).
A conference version of this work has been presented to the 2026 European Signal Processing Conference (EUSIPCO), Bruges, Belgium, Aug. 2026~\cite{lari2026partial}.
The associate editor coordinating the review of this article and approving it for publication was Prof. Hoi-To Wai.
\textit{(Corresponding author: Ehsan Lari.)}}
\thanks{Ehsan Lari is with the Department of Electronic Systems, Norwegian University of Science and Technology (NTNU), Trondheim, Norway (e-mail: ehsanl@alumni.ntnu.no).}
\thanks{Reza Arablouei is with the Commonwealth Scientific and Industrial Research Organisation (CSIRO), Pullenvale, QLD 4069, Australia (e-mail: reza.arablouei@csiro.au).}
\thanks{Stefan Werner is with the Department of Electronic Systems, Norwegian University of Science and Technology (NTNU), Trondheim, Norway. He is also with the Department of Information and Communications Engineering, Aalto University, Espoo, Finland (e-mail: stefan.werner@ntnu.no).}
}


\maketitle

\begin{abstract}

We propose PRISM-FCP (Partial shaRing and robust calIbration with Statistical Margins for Federated Conformal Prediction), a communication-efficient Byzantine-robust federated conformal prediction framework that uses partial model sharing to mitigate stochastic model-poisoning attacks during training and histogram-based filtering to mitigate adversarial calibration submissions.
Existing robust FCP approaches primarily address adversarial behavior during calibration, leaving training-stage poisoning to separate robust-training mechanisms. PRISM-FCP instead considers the coupling between the two stages.
During training, clients partially share updates by transmitting only $M$ of $D$ parameters per round. Under the stated stochastic attack model, this attenuates the expected energy of each Byzantine client's perturbation contribution to the aggregate by a factor of $M/D$ relative to full sharing. When this benefit outweighs the optimization slowdown caused by partial updates, it can reduce training error and improve interval efficiency.
During calibration, the server uses client-provided characterization vectors to filter suspected Byzantine clients before estimating the conformal quantile from the retained clients.
Experiments on synthetic benchmarks and the UCI Superconductivity and YearPredictionMSD datasets, including Gaussian, ALIE, and sign-flipping training-stage attacks, demonstrate near-nominal empirical coverage and favorable communication--performance tradeoffs in the studied settings.

\end{abstract}


\section{Introduction} \label{sec:intro}

\IEEEPARstart{F}{ederated} learning (FL)~\cite{mcmahan2017communication, yang2019federated, li2020federated, kairouz2021advances, lari2025admm} is a distributed learning paradigm in which a population of clients, such as smartphones, hospitals, or financial institutions, collaboratively train a shared model without exchanging raw data. Over the past few years, FL has progressed from early formulations rooted in distributed optimization~\cite{smith2017federated, wang2020tackling, reddi2021adaptive} to increasingly large-scale and production-oriented implementations~\cite{bonawitz2019towards, wang2021field, charles2024towards}. This paradigm is particularly attractive when data is inherently decentralized and centralized training is unfeasible or impractical due to privacy, regulatory, or logistical constraints~\cite{9062302, 9170559, 9770266}. However, FL also introduces new attack surfaces: the server must aggregate client-provided updates that are often unverifiable, making the learning process vulnerable to corrupted or malicious participants~\cite{lari2024icassp, lari2024analyzing}. As FL moves into high-stakes domains, reliable deployment requires not only accurate point predictions, but also calibrated uncertainty with rigorous guarantees. For instance, in a network of hospitals collaboratively training a diagnostic model, a single compromised institution could poison its updates to induce systematically overconfident predictions, leading to unsafe clinical decisions that should have been flagged as uncertain.

Conformal prediction (CP)~\cite{vovk2005algorithmic, plassier2024efficient, 10416237, ye2024benchmarking, zhou2025conformal} is a principled framework for uncertainty quantification that constructs prediction sets (or intervals) with user-specified coverage, assuming only that samples are exchangeable. 
In regression, the distribution-free, finite-sample guarantees established in~\cite{lei2018distribution} provide a canonical foundation for prediction interval construction. In addition, extensions to robust conformal inference can maintain validity under distribution shift, including covariate shift~\cite{ai2024conformal}, a property that is especially relevant in federated environments where client distributions may differ. To bring these benefits to decentralized learning, federated conformal prediction (FCP) has emerged~\cite{lu2023federated, humbert2023one, plassier2023conformal, min2026personalized, koutsoubis2025privacy}, enabling clients to compute prediction intervals without sharing calibration data and addressing practical constraints such as communication efficiency and heterogeneity across clients (e.g., label shift).

In network environments, Byzantine clients pose a critical challenge due to arbitrary and potentially malicious behavior that can derail collaborative training. The work in~\cite{blanchard2017machine} formalizes this threat model and demonstrates that even a single Byzantine participant can drive gradient-based learning toward arbitrarily poor solutions. In practice, adversaries may inject misleading gradients or manipulate model updates~\cite{guerraoui2018hidden, fang2020local}, degrading performance and compromising the integrity of the global model. More recently, self-driven entropy aggregation has been proposed for Byzantine-robust FL under client heterogeneity~\cite{huang2024entropy}, reflecting the realistic regime in which benign clients themselves hold non-IID data. For FCP, the threat is amplified as Byzantine behavior can target not only training but also calibration. Training-phase attacks reduce predictive accuracy, whereas calibration-phase attacks can silently violate coverage guarantees by distorting the estimated conformal quantile, producing overly narrow intervals and systematic undercoverage, a particularly dangerous failure mode in safety-critical applications such as clinical decision support or financial risk assessment.

To mitigate Byzantine behavior during training, a large body of work has proposed robust aggregation rules~\cite{yin2018byzantine, blanchard2017machine, 9721118, tian2022comprehensive}. Coordinate-wise median and trimmed-mean estimators, for example, can suppress outlying (potentially adversarial) updates~\cite{yin2018byzantine}. Recent advances further include Byzantine client identification with false discovery rate control~\cite{qian2024bymi} and privacy-preserving Byzantine-robust mechanisms~\cite{zhang2025practical}. However, these approaches are typically most effective when many clients participate in each round (often under assumptions akin to full participation) and can suffer noticeable degradation under strong client heterogeneity and partial participation, conditions that are common in practical FL deployments.

In the context of FCP, the Rob-FCP algorithm~\cite{kang2024certifiably} addresses calibration-stage attacks by encoding local nonconformity-score distributions into characterization vectors and excluding clients whose vectors deviate significantly from the population, thereby providing certifiable coverage guarantees. Nevertheless, Rob-FCP assumes that training is secure and focuses exclusively on calibration. Conversely, robust training aggregators protect the training phase but remain oblivious to calibration-phase manipulation. 
As a result, existing defenses are \emph{stage-specific} and do not provide an end-to-end solution for Byzantine-robust FCP. This gap motivates methods that can jointly defend both training and calibration while remaining communication-efficient.

Byzantine-robust training and robust conformal calibration can be implemented as distinct modules. However, their guarantees demonstrate an interdependence in the federated CP framework. The nonconformity scores used during calibration are computed using the client predictor obtained after federated training. Hence, a Byzantine perturbation during training affects not only the model error but also the distribution of residual scores observed by benign clients. Poorly trained or unstable client predictors can make benign residual distributions wider and more heterogeneous across clients, causing their histogram characterization vectors to spread in feature space. This reduces benign--Byzantine separability and weakens distance-based calibration filtering. Conversely, when a robust training method yields smaller benign predictor error, the resulting bounds imply tighter concentration around the benign reference distribution and a larger guaranteed calibration-separation margin; empirically, this is accompanied by smaller quantile deviations and narrower intervals in the studied settings. Therefore, robust training and robust calibration are not independent from the viewpoint of coverage, interval width, and filtering reliability.

In this paper, we propose PRISM-FCP, which integrates partial-sharing online federated learning (PSO-Fed)~\cite{9746228, lari2024analyzing} with FCP to mitigate Byzantine effects across both training and calibration phases while reducing communication. In each round, every selected client exchanges only a randomly chosen subset of $M$ out of $D$ model parameters with the server. 
Under the stated stochastic attack model, the random parameter mask acts as a stochastic filter: when a Byzantine client injects an adversarial perturbation, the expected energy of each Byzantine client's perturbation contribution to the aggregate is attenuated by a factor of $M/D$ relative to full sharing. This reduces the injected attack energy; when the resulting attenuation outweighs the optimization slowdown caused by partial updates, it can reduce training error, concentrate benign nonconformity-score distributions, increase their separation from Byzantine calibration submissions, and improve calibration filtering and interval efficiency.

In summary, our main contributions are:
\begin{itemize}
\item \emph{Training-stage robustness via partial sharing:} We show that, under the stated stochastic perturbation model, partial model sharing attenuates the expected per-iteration energy of each Byzantine client's perturbation contribution by the factor $M/D$ relative to full sharing. We then relate steady-state predictor error to residual-quantile and interval-width perturbation bounds (Section~\ref{sec:theory}).
\item \emph{Calibration robustness via improved separability:} We show that a smaller benign training-error bound shrinks the benign histogram concentration radius and, for a fixed adversarial separation profile, increases the guaranteed separation margin used for distance-based Byzantine filtering (Section~\ref{sec:calibration}).
\item \emph{End-to-end empirical validation:} We evaluate PRISM-FCP on synthetic benchmarks and two real-world UCI datasets, Superconductivity and YearPredictionMSD, across different model scales and attack settings. The larger nonlinear YearPredictionMSD experiment, together with Gaussian, ALIE, and sign-flipping attacks, shows that PRISM-FCP maintains near-nominal empirical coverage while providing favorable communication--performance tradeoffs beyond the linear setting considered in the theory (Section~\ref{sec:simulations}).
\end{itemize}

Although PRISM-FCP builds on partial-sharing training and robust histogram-based calibration, the central contribution is the joint training--calibration treatment of Byzantine-robust federated conformal prediction. Our analysis formalizes this coupling by propagating training-stage perturbation bounds into residual-quantile, interval-width, and filtering-reliability bounds.

\textit{Mathematical Notations:} We denote scalars by italic letters, column vectors by bold lowercase letters, and matrices by bold uppercase letters. The superscripts $(\cdot)^{\intercal}$ and $(\cdot)^{-1}$ denote the transpose and inverse operations, respectively, and $\norm{\cdot}$ denotes the Euclidean norm. 
We use $\mathds{1} \{ \cdot \}$ as the indicator function of its event argument and $\Ibf_D$ for the $D \times D$ identity matrix. 
Lastly, $\tr(\cdot)$ denotes the trace of a matrix, and calligraphic letters such as $\Scal$ and $\Bcal$ denote sets.

\section{Preliminaries} \label{sec:pre}

In this section, we introduce the considered system model, review the partial sharing mechanism used in PSO-Fed~\cite{9746228, 9933811, lari2024icassp, lari2024analyzing}, and outline CP in a federated setting.

\subsection{System Model}

We consider a federated network consisting of $K$ clients communicating with a central server. At iteration $n$, client $k$ observes a data (feature-label) pair $\xbf_{k, n} \in \Rbb^D$ and $y_{k, n} \in \Rbb$, generated according to
\begin{align} \label{eq1}
    y_{k, n} = \wbf^{\star\intercal} \xbf_{k, n} + \nu_{k, n},
\end{align}
where $\wbf^\star \in \Rbb^D$ is the unknown parameter vector to be estimated collaboratively and $\nu_{k, n}$ denotes observation noise.
The global learning objective is to minimize the MSE aggregated across clients:
\begin{align} \label{eq2}
    \Jcal(\wbf) = \frac{1}{K} \sum\limits_{k=1}^{K} \Jcal_k(\wbf),
\end{align}
where the local risk at client $k$ is
\[
\Jcal_k(\wbf)=\Ebb\!\left[|y_{k,n}-\wbf^\intercal\xbf_{k,n}|^2\right].
\]
The expectation is taken with respect to the (possibly client-dependent) data distribution at client $k$.
In general, these client distributions may differ substantially (i.e., the data can be non-IID across clients).
The FL goal is to obtain the minimizer $\arg\min_{\wbf} \Jcal(\wbf)$ via decentralized collaboration, without exchanging raw data.

\begin{remark}[Scope of the linear model]
The linear regression model in~\eqref{eq1} is adopted for analytical tractability, enabling a closed-form characterization of Byzantine perturbation attenuation under partial sharing. While a complete theory for more general models is beyond the scope of this work, the underlying mechanism (attenuating the expected energy of each Byzantine client's perturbation contribution by a factor of $M/D$ relative to full sharing) extends conceptually beyond linear regression, and developing such analyses is an interesting direction for future work.
\end{remark}

\subsection{Partial Model Sharing}

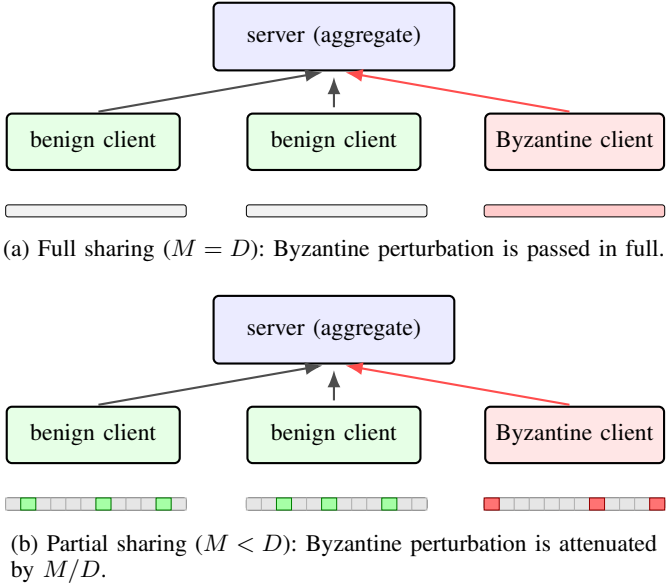
\begin{figure}
\centering
\begin{tikzpicture}[
  font=\footnotesize,
  >=Latex,
  benign/.style={draw, thick, rounded corners=2pt, fill=green!10, inner sep=4pt},
  byz/.style={draw, thick, rounded corners=2pt, fill=red!10, inner sep=4pt},
  server/.style={draw, thick, rounded corners=2pt, fill=blue!8, inner sep=4pt},
  vect/.style={draw, very thin, rounded corners=1pt, fill=gray!10},
  mask/.style={fill=black!10, draw=black!30, very thin},
  goodarrow/.style={-Latex, line width=0.8pt, draw=black!70, shorten >=2pt, shorten <=2pt},
  badarrow/.style={-Latex, line width=0.8pt, draw=red!70, shorten >=2pt, shorten <=2pt},
]

\def\barW{2.4}
\def\barH{0.16}
\def\vbelow{0.45}
\def\nslots{12}
\def\cy{0.9}

\node[server, minimum width=3.2cm, minimum height=0.9cm] (serverA) at (0,2.3) {Aggregation Server};
\node[benign, minimum width=2.3cm, minimum height=0.75cm] (c1A) at (-3.2,\cy) {benign client};
\node[benign, minimum width=2.3cm, minimum height=0.75cm] (c2A) at (-0.0,\cy) {benign client};
\node[byz,    minimum width=2.3cm, minimum height=0.75cm] (c3A) at ( 3.2,\cy) {Byzantine client};

\draw[vect] (c1A.south west) ++(0.0,-\vbelow) rectangle ++(\barW,-\barH);
\draw[vect] (c2A.south west) ++(0.0,-\vbelow) rectangle ++(\barW,-\barH);
\draw[vect, fill=red!20] (c3A.south west) ++(0.0,-\vbelow) rectangle ++(\barW,-\barH);

\draw[goodarrow] (c1A.north) -- ($(serverA.south west)!0.95!(serverA.south)$);
\draw[goodarrow] (c2A.north) -- (serverA.south);
\draw[badarrow]  (c3A.north) -- ($(serverA.south east)!0.95!(serverA.south)$);

\node[align=center] at (0,-0.55) {(a) Full sharing ($M=D$): Byzantine perturbation is passed in full.};

\begin{scope}[yshift=-3.9cm] 
\node[server, minimum width=3.2cm, minimum height=0.9cm] (serverB) at (0,2.3) {Aggregation Server};
\node[benign, minimum width=2.3cm, minimum height=0.75cm] (c1B) at (-3.2,\cy) {benign client};
\node[benign, minimum width=2.3cm, minimum height=0.75cm] (c2B) at (-0.0,\cy) {benign client};
\node[byz,    minimum width=2.3cm, minimum height=0.75cm] (c3B) at ( 3.2,\cy) {Byzantine client};

\draw[vect] (c1B.south west) ++(0.0,-\vbelow) rectangle ++(\barW,-\barH);
\draw[vect] (c2B.south west) ++(0.0,-\vbelow) rectangle ++(\barW,-\barH);
\draw[vect, fill=red!20] (c3B.south west) ++(0.0,-\vbelow) rectangle ++(\barW,-\barH);

\foreach \i in {0,...,11} {
  \draw[mask] (c1B.south west) ++(0.0 + \i*\barW/\nslots,-\vbelow) rectangle ++(\barW/\nslots,-\barH);
  \draw[mask] (c2B.south west) ++(0.0 + \i*\barW/\nslots,-\vbelow) rectangle ++(\barW/\nslots,-\barH);
  \draw[mask] (c3B.south west) ++(0.0 + \i*\barW/\nslots,-\vbelow) rectangle ++(\barW/\nslots,-\barH);
}

\def\slotsCOne{1,6,10}
\def\slotsCTwo{2,5,9}
\def\slotsCThree{0,7,11}
\foreach \x in \slotsCOne {\draw[fill=green!35, draw=green!50!black, very thin]
    (c1B.south west) ++(0.0 + \x*\barW/\nslots,-\vbelow)
    rectangle ++(\barW/\nslots,-\barH);}
\foreach \x in \slotsCTwo {\draw[fill=green!35, draw=green!50!black, very thin]
    (c2B.south west) ++(0.0 + \x*\barW/\nslots,-\vbelow)
    rectangle ++(\barW/\nslots,-\barH);}

\foreach \x in \slotsCThree {\draw[fill=red!55, draw=red!60!black, very thin]
    (c3B.south west) ++(0.0 + \x*\barW/\nslots,-\vbelow)
    rectangle ++(\barW/\nslots,-\barH);}

\draw[goodarrow] (c1B.north) -- ($(serverB.south west)!0.95!(serverB.south)$);
\draw[goodarrow] (c2B.north) -- (serverB.south);
\draw[badarrow]  (c3B.north) -- ($(serverB.south east)!0.95!(serverB.south)$);

\node[align=left] at (0,-0.75) {(b) Partial sharing ($M<D$): the expected energy of each Byzantine client's\\ perturbation contribution is attenuated by $M/D$ relative to full sharing.};
\end{scope}

\end{tikzpicture}

\caption{Illustration of how partial sharing attenuates Byzantine perturbations. 
}
\label{fig:ps-attenuation}
\end{figure}

In our prior work~\cite{lari2024icassp, lari2024analyzing}, we have studied partial model sharing through PSO-Fed in detail (see Fig.~\ref{fig:ps-attenuation}). Here, we summarize the mechanism most relevant to this paper. Unlike conventional FL, where the full parameter vector is exchanged between the server and clients at every iteration, PSO-Fed communicates only a subset of parameters. Specifically, at iteration $n$, client $k$ receives the masked global model estimate and uploads only a masked version of its local model estimate.
This masking is represented by a diagonal selection matrix $\Sbf_{k, n} \in \Rbb^{D \times D}$ with exactly $M$ ones on the diagonal (and $D-M$ zeros), which specifies the model parameters communicated between client $k$ and the server at iteration $n$. Therefore, the PSO-Fed recursions for minimizing \eqref{eq2} are given by
\begin{align*}
\epsilon_{k,n}
&= y_{k,n}-\Bigl[\Sbf_{k,n-1}\wbf_{n-1} \\
&\quad +(\Ibf_D-\Sbf_{k,n-1})\wbf_{k,n-1}\Bigr]^\intercal\xbf_{k,n}, \\
\wbf_{k,n}
&= \Sbf_{k,n-1}\wbf_{n-1}+(\Ibf_D-\Sbf_{k,n-1})\wbf_{k,n-1} \\
&\quad +\mu\xbf_{k,n}\epsilon_{k,n}, \\
\wbf_n
&=\frac{1}{|\Scal_n|}\sum_{k\in\Scal_n}\Sbf_{k,n}\wbf_{k,n} \\
&\quad+\frac{1}{|\Scal_n|}\sum_{k\in\Scal_n}(\Ibf_D-\Sbf_{k,n})\wbf_{n-1}.
\end{align*}
where $\wbf_{k, n}$ is the local model estimate at client $k$ and iteration $n$, $\wbf_n$ is the global model estimate at iteration $n$, $\mu$ is the stepsize parameter, and $\Scal_n$ is the set of participating clients at iteration $n$, and $|\Scal_n|$ is the cardinality of $\Scal_n$.

\subsection{Byzantine Attack during Training Phase}\label{sec:training_attack}

\begin{figure*}[t!]
\centering
\subfloat[Efficiency attack\label{fig:conv-vs-nrobustADMM}]{{\includegraphics[width=.3333\textwidth]{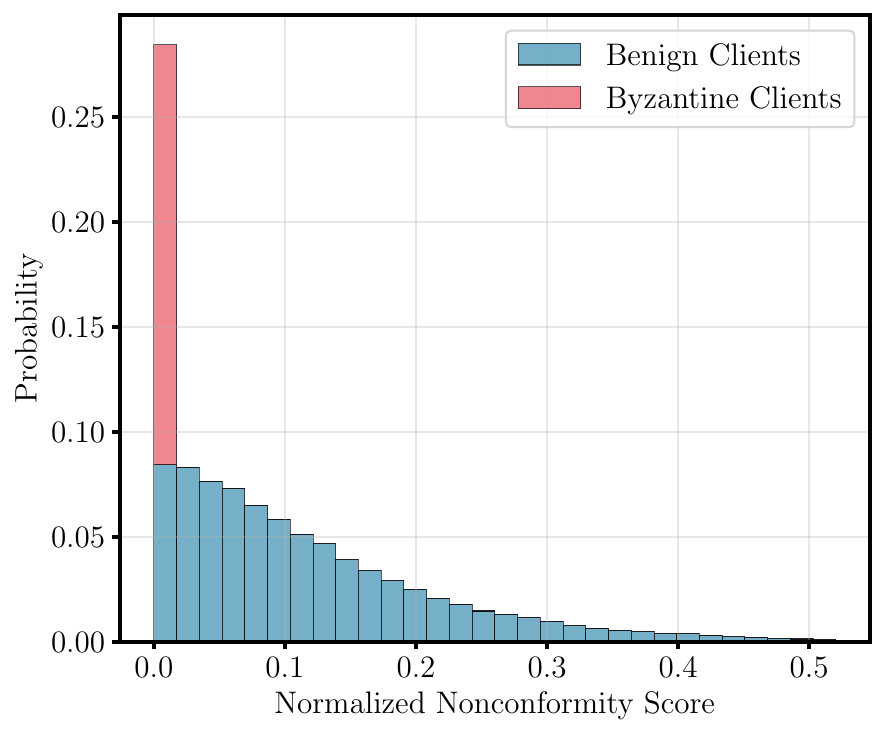}}}
\hfill
\subfloat[Coverage attack\label{fig:conv-vs-nrobustADMM_diff}]{{\includegraphics[width=.3333\textwidth]{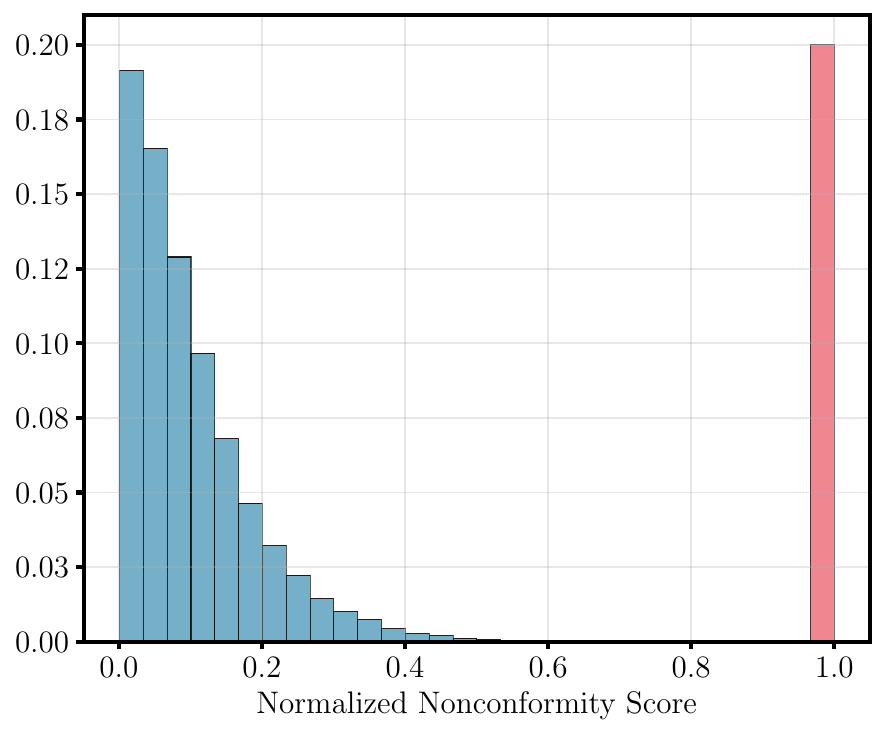}}}
\hfill
\subfloat[Random attack\label{fig:prop_nrADMM_diff}]{{\includegraphics[width=.3333\textwidth]{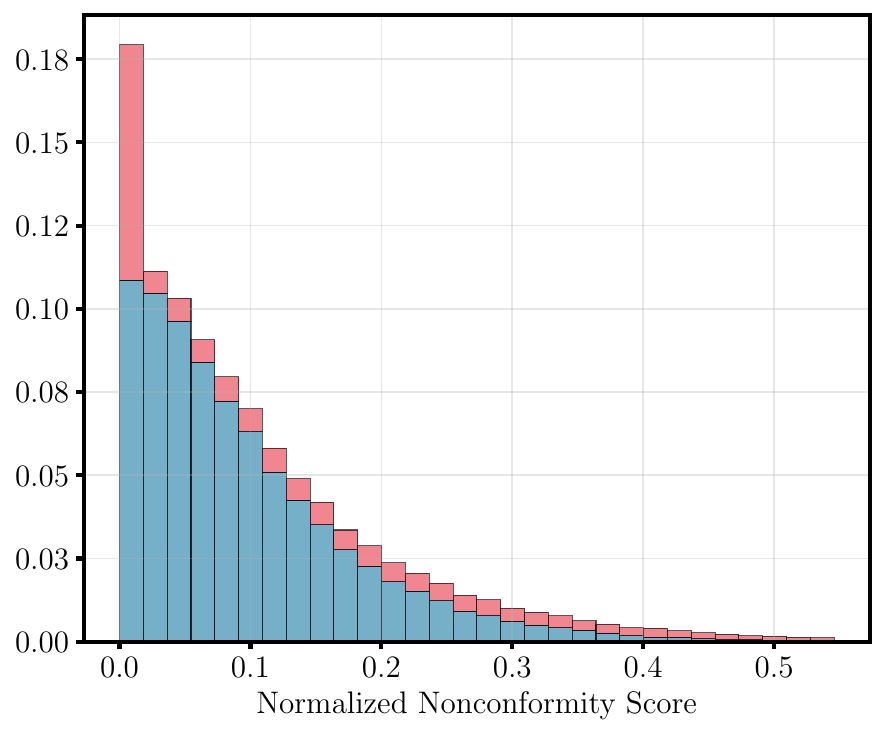}}}
\caption{Histograms illustrating the effect of different Byzantine attacks during the calibration phase: (a) efficiency attack (adversaries report all-zero normalized scores), (b) coverage attack (adversaries report all-one normalized scores), and (c) random attack (adversaries add Gaussian noise to their scores). 
}
\label{fig:DiffAttacks}
\end{figure*}

Let $\Scal_B$ denote the set of Byzantine clients, and let $\beta_k\in\{0,1\}$ indicate whether client $k$ is Byzantine ($\beta_k = 1$ if client $k \in \Scal_B$ and $\beta_k = 0$ otherwise). The number of Byzantine clients is $|\Scal_B|$, which is assumed to be known by the server\footnote{This assumption can be relaxed to an upper bound or an estimate.}.
We consider an uplink attack model in which, at each iteration, every Byzantine client perturbs its transmitted (partial) model update with probability $p_a$. Specifically, we model the corruption as $\wbf_{k,n}+\tau_{k,n}\boldsymbol{\delta}_{k,n}$, where $\tau_{k,n}$ is a Bernoulli random variable with $\Pr(\tau_{k,n}=1)=p_a$, and $\boldsymbol\delta_{k,n}\sim\Ncal(\mathbf{0},\sigma_B^2\Ibf_D)$ is zero-mean white Gaussian noise~\cite{7563348}. Accordingly, instead of the benign partial update $\Sbf_{k,n}\wbf_{k,n}$, the server receives
\[
\Sbf_{k,n}\wbf_{k,n}
+\beta_k\tau_{k,n}\Sbf_{k,n}\boldsymbol{\delta}_{k,n}.
\]
This model captures stochastic Byzantine behavior that injects additive noise into the aggregation process. 

\begin{remark}[Stochastic vs.\ adversarial attack model]
The Gaussian perturbation model is standard in Byzantine-robust distributed learning~\cite{7563348, blanchard2017machine} and enables closed-form analysis. More powerful adversaries may adapt their perturbations to the shared coordinates if they observe the selection masks $\Sbf_{k,n}$. In our setting, however, the Byzantine perturbations are independent of the selection masks (Assumption A2 in Section~\ref{sec:asdef}), while the masks are randomized independently across clients and iterations (Assumption A3). Under these assumptions, partial sharing attenuates the expected energy of each Byzantine client's perturbation contribution by a factor of $M/D$ relative to full sharing. Extending the analysis and guarantees to fully adaptive, mask-aware adversaries is an interesting direction for future work.
\end{remark}

\subsection{Federated Conformal Prediction}

CP for regression constructs prediction intervals with a user-specified confidence level (e.g., $90\%$) under the assumption of data exchangeability. Given a trained predictor, CP calculates nonconformity scores (typically absolute residuals) on a calibration set and uses an empirical quantile of these scores to form prediction intervals with distribution-free marginal coverage. Importantly, these guarantees do not rely on the correctness of the underlying model and hold for arbitrary model classes.

In the standard CP pipeline, we train a model on a training set and compute conformal calibration scores on a held-out calibration dataset $\{(\xbf_j,y_j)\}_{j=1}^N$ using the learned model parameter vector $\hat{\wbf}$. The nonconformity score for each calibration sample is $r_j = |y_j - \hat{\wbf}^\intercal \xbf_j|$, $j = 1, \dots, N$. Let $q_{1-\alpha}$ be the $\lceil (N+1)(1-\alpha) \rceil$-th order statistic of $\{r_j\}_{j=1}^{N}$.
Then, the two-sided $(1 - \alpha)$ prediction interval for a new input $\xbf$ is
\begin{align*}
    \Ccal (\xbf) = \left[ \hat{\wbf}^\intercal \xbf - q_{1 - \alpha},\; \hat{\wbf}^\intercal \xbf + q_{1 - \alpha} \right].
\end{align*}
The confidence parameter $\alpha\in (0, 1)$ controls interval width: smaller $\alpha$ yields wider intervals. Under exchangeability, CP guarantees marginal coverage
\[
\Pbb\!\left(y\in\Ccal(\xbf)\right)\ge 1-\alpha.
\]

In FL, after training of a global model $\hat{\wbf}$ and local models $\hat{\wbf}_k$, each client $k$ holds a local calibration set $\{ ( \xbf_{k,j}, y_{k,j} ) \}_{j=1}^{N_k}$ (with \emph{partial exchangeability}~\cite{de1980partial, lu2023federated} assumption) and computes local nonconformity scores
\[
r_{k,j}=|y_{k,j}-\hat{\wbf}_k^\intercal\xbf_{k,j}|,
\qquad j=1,\ldots,N_k.
\]
Privacy constraints prevent clients from sharing raw calibration data (or even individual scores) with the server. The FCP algorithm~\cite{lu2023federated} addresses this by employing secure aggregation or sketching mechanisms (e.g., T-Digest~\cite{dunning2021t}) to enable the server to approximate the global conformal quantile from the pooled score population as
\begin{align}
    \hat{q}_{1 - \alpha} \approx \mathrm{Quantile}_{1 - \alpha} \left( \bigcup_{k=1}^K \{ r_{k,j} \}_{j=1}^{N_k} \right).
\label{eq:quantile}
\end{align}
The server then broadcasts $\hat{q}_{1-\alpha}$, and each client combines this common quantile with its local predictor $\hat{\wbf}_k$ to construct its local prediction interval. In practice, when the sketching/aggregation error is small, this approximation can preserve coverage closely while protecting client-level calibration privacy.

\subsection{Byzantine Attack during Calibration Phase}

During the calibration phase, Byzantine clients in $\Scal_B$ may submit arbitrary or adversarially crafted nonconformity scores (or score summaries) instead of their true values, with the aim of biasing the global quantile estimate used to construct prediction intervals. By manipulating the estimated quantile $q_{1-\alpha}$, adversaries can induce either overly narrow intervals (causing systematic undercoverage) or overly wide intervals (yielding vacuous uncertainty), thereby compromising the practical reliability of CP.

In this work, we consider three representative calibration-phase Byzantine attacks, whose effects are illustrated in Fig.~\ref{fig:DiffAttacks}: 
\begin{itemize}[leftmargin=*]
\item \emph{Efficiency attack}: Byzantine clients report near-zero scores (e.g., all zeros after normalization), deflating $q_{1-\alpha}$ and leading to undercoverage.
\item \emph{Coverage attack}: Byzantine clients report maximal scores (e.g., all ones after normalization), inflating $q_{1 - \alpha}$ and producing excessively wide prediction intervals. 
\item \emph{Random attack}: Byzantine clients perturb their scores by additive noise, injecting randomness into the quantile estimation (e.g., $\tilde r_{k,j}=r_{k,j}+\eta_{k,j}$ with $\eta_{k,j}\sim\Ncal(0,\sigma_C^2)$, optionally clipped to the score range).
\end{itemize}

\section{Byzantine-Robust FCP via Partial Sharing} \label{sec:method}

In this section, we present PRISM-FCP, an efficient FCP framework that mitigates Byzantine attacks in both the training and calibration phases. The key insight is that the partial model sharing mechanism~\cite{lari2024analyzing}, which is originally designed for communication efficiency, also attenuates Byzantine perturbations during training. When this attenuation yields a smaller benign training error, it can improve downstream calibration robustness by making benign score distributions more concentrated and easier to separate from outliers.

\subsection{Mitigating Byzantine Attack during Training}\label{sec:train-mitig}

Partial model sharing reduces communication overhead by allowing each participating client to exchange only a subset of model parameters with the server~\cite{lari2024icassp, lari2024analyzing}. Beyond communication savings, partial sharing provides an attenuation mechanism against the considered stochastic model-poisoning attacks. In particular, prior results establish mean and mean-square convergence of PSO-Fed under noisy Byzantine updates~\cite{lari2024analyzing}. In this work, we exploit this mechanism to mitigate Byzantine influence during training; when the attenuation benefit dominates the slower progress caused by partial updates, the resulting lower training error also benefits conformal calibration.

\subsection{Mitigating Byzantine Attacks during Calibration}\label{sec:calib-mitig}

To mitigate Byzantine attacks in the calibration phase, each benign client computes nonconformity scores $\{ r_{k,j} \}_{j=1}^{N_k}$ on its local calibration set and summarizes them via a histogram-based characterization vector $\vbf_k \in \Delta^H$. Let $0=a_0<a_1<\cdots<a_H=1$ be fixed bin edges and define
\[
I_h=[a_{h-1},a_h),\quad h=1,\ldots,H-1,\qquad I_H=[a_{H-1},1].
\]
Then
\begin{align*}
    v_{k,h} = \frac{1}{N_k}\sum_{j=1}^{N_k} \mathds{1}\!\{\tilde r_{k,j}\in I_h\}, \quad h=1,\dots,H,
\end{align*}
where $\tilde r_{k,j}$ denotes a normalized (or clipped) score mapped to $[0,1]$. Thus, the bins form a partition of $[0,1]$, including the clipped value $1$. Each entry $v_{k,h}$ is the empirical probability that the local scores fall into bin $h$. The set 
\[\Delta^H \triangleq \{ \vbf_k \in \mathbb{R}^H | v_{k,h} \geq 0, \sum_{h=1}^H v_{k,h} = 1 \}\]
denotes the $(H-1)$-dimensional probability simplex. Clients transmit $\vbf_k$ to the server, while Byzantine clients may submit arbitrary or adversarial vectors (cf.\ Fig.~\ref{fig:DiffAttacks} with $H = 30$). 

\begin{remark}[Normalization of scores]
Before histogram construction, residual scores are normalized to a common range. In our experiments, we determine this range using a pilot calibration pass. Let $r_{k,j}$ denote the absolute residual of sample $j$ at client $k$. We use
\[
\begin{aligned}
R_{\max}&=1.1\max_{k,j}r_{k,j},\\
\tilde r_{k,j}&=\frac{\min\{r_{k,j},R_{\max}\}}{R_{\max}}.
\end{aligned}
\]
The histogram is then constructed over the normalized interval $[0,1]$. The factor $1.1$ provides a small margin beyond the largest observed pilot residual. For the finite-sample concentration analysis in Section~\ref{sec:calibration}, the normalization range and bin edges are treated as fixed and are assumed to be selected independently of the calibration samples used to form $\vbf_k$. A separate pilot set is one way to satisfy this condition. If $R_{\max}$ is chosen too large, the bins become wide relative to the residual distribution and may lose discriminative power; if it is too small, excessive clipping may obscure distributional differences. In deployments where the pilot range may itself be affected by adversarial clients, robust alternatives include using a high empirical quantile, a median/MAD-based range estimate, or an overflow bin for clipped residuals.
\end{remark}

This histogram summary preserves coarse distributional structure without transmitting raw scores, reducing the granularity of the information revealed to the server. Moreover, benign clients tend to produce similar characterization vectors, whereas Byzantine submissions typically deviate, enabling robust detection.

\begin{remark}[Privacy in histogram binning]
The histogram characterization vector transmits bin counts rather than the raw calibration scores, which reduces the granularity of the calibration information revealed to the server. However, histogram binning alone does not provide a formal privacy guarantee. In particular, the histogram may still reveal distributional information about a client's calibration residuals. Thus, the proposed binning step should be interpreted as an information-reduction mechanism, not as a Differential Privacy (DP)~\cite{dwork2014algorithmic} or cryptographic privacy mechanism. If formal privacy guarantees are required, Differential Privacy mechanisms such as adding calibrated noise to the histogram counts can be incorporated into the PRISM-FCP calibration pipeline, at the cost of introducing an additional privacy--robustness--accuracy tradeoff.
\end{remark}

We compute pairwise $\ell_2$ distances between characterization vectors as 
\[d_{k, k^{\prime}} = \left\lVert \vbf_k - \vbf_{k^{\prime}} \right\rVert_2,\ \forall\, k, k^{\prime} \in \{1, \cdots, K\}.\]
Let $K_b := K-|\Scal_B|$ denote the number of benign clients. Following Rob-FCP~\cite{kang2024certifiably}, we assign each client $k$ a maliciousness score, $m_k$, as the sum of distances to its $K_b - 1$ farthest neighbors:
\begin{align}\label{eq:mal_score}
    m_k = \sum_{k^{\prime} \in \mathrm{Far}(k, K_b-1)} d_{k,k^{\prime}},
\end{align}
where $\mathrm{Far}(k, K_b-1)$ is the set of the $K_b - 1$ other clients with the largest distances from client $k$. Intuitively, Byzantine clients lie farther from the benign population and hence attain larger $m_k$. We declare the $|\Scal_B|$ clients with the largest maliciousness scores as Byzantine and form the benign set $\Bcal$ from the remaining clients.

Finally, after filtering, the server estimates the conformal quantile $\hat{q}_{1 - \alpha}$ using only secure score sketches, quantile summaries, or otherwise privacy-preserving score summaries contributed by clients in $\Bcal$, and broadcasts it to the clients. Each client then combines this common quantile with its local predictor to construct its prediction interval. This calibration-stage filtering follows Rob-FCP's robust detection procedure~\cite{kang2024certifiably}, while PRISM-FCP additionally attenuates training-stage Byzantine influence and reduces communication via partial sharing. We summarize the complete end-to-end pipeline in Algorithm~\ref{alg:prism-fcp}.

\setlength{\algomargin}{0.5em}
\begin{algorithm}[t!]
\caption{PRISM-FCP Workflow.}
\label{alg:prism-fcp}
\small
\DontPrintSemicolon
\SetAlgoLined
\KwIn{dimension $D$; sharing $M$; miscoverage level $\alpha$; bins $H$; stepsize $\mu$; normalization range $R_{\max}$}
\KwOut{Client-specific prediction intervals $\{\Ccal_k(\xbf)\}_{k=1}^K$ for test input $\xbf$}
\BlankLine
\tcc{\hspace{1.3cm}\textbf{Phase 1: Robust Training}}
Initialize $\wbf_0 \gets \mathbf{0}$, $\wbf_{k,0} \gets \mathbf{0}$, and client-specific selection matrices $\Sbf_{k,0}$ with $\mathrm{tr}(\Sbf_{k,0})=M$ for all $k$\;
\For{$n = 1, 2, \ldots, N$}{
    Server selects participant set $\Scal_n \subseteq \{1, \ldots, K\}$\;
    Server generates client-specific selection matrices $\{\Sbf_{k,n}\}_{k=1}^K$ with $\mathrm{tr}(\Sbf_{k,n})=M$\;
    \For(\tcp*[f]{in parallel}){each $k \in \Scal_n$}{
        Receive masked global model $\Sbf_{k,n-1} \wbf_{n-1}$\;
        Observe $(\xbf_{k,n}, y_{k,n})$\;
        $\bar{\wbf}_{k,n} \gets \Sbf_{k,n-1} \wbf_{n-1} + (\Ibf_D - \Sbf_{k,n-1}) \wbf_{k,n-1}$\;
        $\epsilon_{k,n} \gets y_{k,n} - \bar{\wbf}_{k,n}^\intercal \xbf_{k,n}$\;
        $\wbf_{k,n} \gets \bar{\wbf}_{k,n} + \mu \, \xbf_{k,n} \, \epsilon_{k,n}$\;
        Upload $\ubf_{k,n} \gets \Sbf_{k,n} \wbf_{k,n}$\;
        \tcp*[l]{Byzantine: $\ubf_{k,n} \mathrel{+}= \tau_{k,n} \Sbf_{k,n} \boldsymbol{\delta}_{k,n}$}
    }
    \For{each $k \notin \Scal_n$}{
        Local-only update using $(\xbf_{k,n}, y_{k,n})$\;
    }
    $\wbf_n \gets \frac{1}{|\Scal_n|} \sum_{k \in \Scal_n} \left[\ubf_{k,n} + (\Ibf_D-\Sbf_{k,n})\wbf_{n-1}\right]$\;
}
$\hat{\wbf} \gets \wbf_N$, \quad $\hat{\wbf}_k \gets \wbf_{k,N}$ for all $k$\;
\BlankLine
\tcc{\hspace{1.1cm}\textbf{Phase 2: Robust Calibration}}
\For(\tcp*[f]{in parallel}){each $k = 1, \ldots, K$}{
    $r_{k,j} \gets |y_{k,j} - \hat{\wbf}_k^\intercal \xbf_{k,j}|$, \quad $j = 1, \ldots, N_k$\;
    $\tilde{r}_{k,j} \gets \min\{r_{k,j},R_{\max}\}/R_{\max}$\;
    $v_{k,h} \gets \frac{1}{N_k} \sum_{j=1}^{N_k} \mathds{1}\{\tilde{r}_{k,j}\in I_h\}$, \quad $h = 1, \ldots, H$\;
    Construct a score sketch or quantile summary $\mathcal Q_k$ from $\{r_{k,j}\}_{j=1}^{N_k}$, including its sample count $N_k$\;
    Upload $\vbf_k = [v_{k,1}, \ldots, v_{k,H}]^\intercal$ and $\mathcal Q_k$ to server\;
    \tcp*[l]{\footnotesize Byzantines upload adversarial summaries}
}
\uIf{$|\Scal_B|$ is known by the server}{
    Compute $d_{k,k'} = \|\vbf_k - \vbf_{k'}\|_2$ for all $k, k'$\;
    $m_k \gets \sum_{k' \in \mathrm{Far}(k,\, K - |\Scal_B| - 1)} d_{k,k'}$\;
    $\Bcal \gets \{1, \ldots, K\} \setminus \{\text{top-}|\Scal_B| \text{ clients by } m_k\}$\;
}
\Else{
    $\vbf_{\mathrm{med}} \gets \mathrm{CoordinateWiseMedian}(\{\vbf_k\}_{k=1}^K)$\;
    $m_k \gets \|\vbf_k - \vbf_{\mathrm{med}}\|_2$ for all $k$\;
    $\Bcal \gets \{k \mid m_k \leq \text{MAD threshold}\}$; 
}
Use the stored score sketches or quantile summaries $\mathcal Q_k$ from retained clients $k\in\Bcal$\;
$\hat{q}_{1-\alpha} \gets \mathrm{FCPQuantile}_{1-\alpha}\big(\{\mathcal Q_k:k\in\Bcal\}\big)$\;
\BlankLine
\tcc{\hspace{1.2cm}\textbf{Phase 3: Testing/Inference}}
\For{each $k = 1, \ldots, K$}{
    $\Ccal_k(\xbf) \gets
    \big[\hat{\wbf}_k^\intercal \xbf - \hat{q}_{1-\alpha},\;
    \hat{\wbf}_k^\intercal \xbf + \hat{q}_{1-\alpha}\big]$\;
}
\end{algorithm}

\begin{remark}[Handling unknown $|\Scal_B|$]~\label{rem:handling_unknown}
When the number of Byzantine clients is unknown, we replace the top-$|\Scal_B|$ filtering with a robust outlier rule such as the median absolute deviation (MAD) criterion~\cite{leys2013mad}. Specifically, we compute each client's distance to a robust population center (e.g., the coordinate-wise median characterization vector) and flag clients whose distance exceeds a MAD-based threshold. Our experiments in Section~\ref{sec:simulations} demonstrate that both known-$|\Scal_B|$ filtering and MAD-based filtering can preserve nominal coverage under attack. 
\end{remark}

\begin{remark}[Scalability of calibration filtering]~\label{rem:scalability_filtering}
The maliciousness score in the calibration stage is computed from pairwise distances between the clients' $H$-dimensional calibration-score histograms. Therefore, a direct implementation requires $\mathcal{O}(K^2H)$ arithmetic operations and $\mathcal{O}(K^2)$ memory if the full distance matrix is stored. This cost is incurred only once after the federated training stage, rather than at every communication round. Hence, for the moderate federation sizes considered in our experiments, the overhead is negligible compared with iterative model training. However, in very large-scale cross-device deployments with thousands or tens of thousands of clients, exact all-pairs scoring may become computationally expensive. In such settings, the calibration filter can be implemented using scalable approximations, such as random pair subsampling, anchor- or prototype-based comparisons, locality-sensitive hashing for approximate nearest-neighbor search, blockwise streaming computation, or clustering-based pre-screening. These approximations reduce the effective number of pairwise comparisons, at the cost of replacing the exact maliciousness score with an approximate one.
\end{remark}

\begin{remark}[Farthest-neighbor scoring]~\label{rem:scoring}
We define the maliciousness score as the sum of distances to the $K_b - 1$ farthest neighbors. This formulation is equivalent, up to a scaling constant, to the average of the farthest-neighbor distances utilized in standard implementations. We adopt the sum formulation as it facilitates a more direct theoretical analysis of the separation bounds.
\end{remark}

\section{Theoretical Analysis} \label{sec:theory}

We analyze PRISM-FCP under the stated stochastic training-stage attack model and calibration-stage separation assumptions. Lemma~\ref{lem:partial-sharing-attenuation} establishes the $M/D$ attenuation of the expected per-iteration energy of each Byzantine client's perturbation contribution. Corollaries~\ref{cor:quantile-ps}--\ref{cor:quantile-exact} and Proposition~\ref{prop:width} relate predictor error to quantile and width deviations. Proposition~\ref{prop:margin} and Theorem~\ref{thm:detection} then connect smaller benign training error to calibration-stage separation and filtering under their stated conditions.

\subsection{Assumptions and Definitions} \label{sec:asdef}

To make the analysis tractable, we adopt the following standard assumptions.

\noindent {A1:} The input vectors $\xbf_{k,n}$ at each client $k$ are generated by a wide-sense stationary random process with finite second-moment matrix $\Rbf_k = \Ebb \big[ \xbf_{k,n}\xbf_{k,n}^\intercal \big]$.

\noindent {A2:} The observation noises $\nu_{k,n}$ are zero-mean and independent across clients and iterations, with finite, possibly client-dependent variances. The attack indicators $\tau_{k,n}$ are IID Bernoulli$(p_a)$, and the Byzantine perturbations $\boldsymbol{\delta}_{k,n}$ are IID, zero-mean, with covariance $\sigma_B^2\Ibf_D$. The observation noises, attack indicators, Byzantine perturbations, input processes, and selection masks are mutually independent whenever they appear in the same update.

\noindent {A3:} Each selection matrix $\Sbf_{k,n}$ contains exactly $M$ ones on its diagonal, chosen uniformly without replacement from the $D$ coordinates, and the selection matrices are independent across clients and iterations.

\subsection{Mean and Mean-Square Convergence and Steady-State Error Measures}

We summarize key results from~\cite{lari2024analyzing} for PSO-Fed and distinguish two error quantities used below: the coefficient mean-square deviation (MSD), which measures model-parameter error, and the output prediction MSE. The convergence results recalled below are invoked under the corresponding moment and stability conditions established in~\cite{lari2024analyzing}, in addition to A1--A3; A1--A3 identify the assumptions used explicitly in the new derivations that follow.

\begin{lemma}[Mean convergence]\label{lem:mean-conv}
PSO-Fed converges in the mean sense under Byzantine attacks with a suitable stepsize~\cite[Sec.~III-A]{lari2024analyzing}.
\end{lemma}

\begin{lemma}[Mean-square convergence]\label{lem:ms-convergence}
PSO-Fed converges in the mean-square sense under Byzantine attacks with a suitable stepsize~\cite[Sec.~III-B]{lari2024analyzing}.
\end{lemma}

\begin{lemma}[Steady-state coefficient MSD and prediction MSE]\label{lem:steady-state-mse}
Under mean-square convergence, the coefficient-error second moment approaches a finite steady-state value; for client $k$, define
\begin{align}\label{eq:steady-state-msd}
\mathcal D_k:=\lim_{n\to\infty}\Ebb\|\wbf_{k,n}-\wbf^\star\|_2^2<\infty.
\end{align}
Separately, the steady-state output prediction MSE derived in~\cite[Sec.~III-C]{lari2024analyzing} admits the decomposition
\begin{align}\label{eq:steady-state-mse}
\Ecal=\Ecal_{\boldsymbol{\phi}}+\Ecal_{\boldsymbol{\omega}}+\Ecal_{\boldsymbol{\Theta}}.
\end{align}
\end{lemma}

\begin{remark}[Interpretation of the error measures]\label{rem:mse-decomposition}
The coefficient MSD $\mathcal D_k$ in~\eqref{eq:steady-state-msd} is the quantity used in the parameter-error bounds below. The output prediction MSE in~\eqref{eq:steady-state-mse} is a distinct quantity: $\Ecal_{\boldsymbol{\phi}}$ captures observation-noise propagation through the learning dynamics, $\Ecal_{\boldsymbol{\omega}}$ is the contribution induced by Byzantine perturbations, and $\Ecal_{\boldsymbol{\Theta}}$ is the irreducible observation-noise term~\cite{lari2024analyzing}. These two error measures should not be identified with one another.
\end{remark}

\begin{lemma}[Partial sharing attenuates the Byzantine driving perturbation]\label{lem:partial-sharing-attenuation}
Under A2--A3, when $\Sbf_{k,n}$ selects $M$ coordinates, the expected per-iteration energy of each Byzantine client's mask-oblivious isotropic perturbation contribution to the aggregate is reduced by the exact factor $M/D$ relative to full sharing. The resulting steady-state quantities $\mathcal D_k$ and $\Ecal_{\boldsymbol{\omega}}$ depend on the $M$-dependent learning dynamics and are not asserted to scale exactly as $M/D$~\cite{lari2024analyzing}.
\end{lemma}

\subsection{Training-Error Consequences for FCP Quantiles} \label{sec:consequences}

We next relate predictor error to conformal calibration. The results below show that a smaller coefficient error tightens bounds on the deviation of the residual quantile and interval width from their benign-reference values; they do not by themselves impose a monotonic direction on the realized quantile or width.

A central objective of CP is to produce intervals that are both \emph{valid} (achieving target marginal coverage $1-\alpha$) and \emph{efficient} (as narrow as possible). Validity is ensured by conformal quantile calibration under exchangeability, whereas efficiency depends on the nonconformity-score distribution, which is influenced by the trained predictor.

For this subsection, let $\mathcal T_n$ denote the sigma-field generated by the training procedure up to iteration $n$. Conditional on $\mathcal T_n$, the trained predictor $\wbf_{k,n}$ is fixed. Let $(\xbf_k^{\mathrm{cal}},y_k^{\mathrm{cal}})$ be a fresh benign calibration draw from client $k$, independent of $\mathcal T_n$. We write $\Ebb_{\mathrm{cal}}[\cdot\mid\mathcal T_n]$ for expectation over this calibration draw conditional on training and $\Ebb_{\mathrm{tr}}[\cdot]$ for expectation over training randomness. This conditioning matches the operational conformal setting in which calibration is performed on held-out data after training.

\begin{lemma}[Lipschitz continuity of residuals]\label{lem:residual-lip}
For a sample $(\xbf,y)$, let $r(\wbf):=|y-\wbf^\intercal\xbf|$. For any $\wbf$ and $\wbf^\star$,
\[
|r(\wbf)-r(\wbf^\star)|\le \|\xbf\|_2\,\|\wbf-\wbf^\star\|_2.
\]
\end{lemma}

\begin{proofe}
Let $a=y-\wbf^\intercal\xbf$ and $b=y-\wbf^{\star\intercal}\xbf$. By the reverse triangle inequality, $\big||a|-|b|\big|\le |a-b|$. Moreover,
\[
|a-b|=\left|(\wbf-\wbf^\star)^\intercal\xbf\right|\le \|\wbf-\wbf^\star\|_2\,\|\xbf\|_2,
\]
by Cauchy--Schwarz.$\hfill\IEEEQED$
\end{proofe}

\begin{remark}[Applicability beyond partial sharing]
The training-to-calibration arguments developed from the residual-stability analysis onward are not specific to partial sharing. In addition to the calibration-stage regularity and separation assumptions stated below, they require a suitable bound on the discrepancy between the predictor obtained after training and a benign reference predictor. Consequently, under those assumptions, the subsequent quantile-stability, histogram-concentration, and filtering-separation arguments can in principle be combined with other Byzantine-robust training schemes that provide a compatible predictor-error bound. Partial sharing is the communication-efficient instantiation analyzed explicitly in this work.
\end{remark}

\begin{theorem}[Quantile stability under local density bounds]\label{thm:quantile-stability}
For a fixed client $k$, define
\[
X_{k,n}:=\big|y_k^{\mathrm{cal}}-\wbf_{k,n}^\intercal\xbf_k^{\mathrm{cal}}\big|,
\qquad
Y_k:=\big|y_k^{\mathrm{cal}}-\wbf^{\star\intercal}\xbf_k^{\mathrm{cal}}\big|,
\]
and let
\[
F_{k,n}(t):=\Pbb(X_{k,n}\le t\mid\mathcal T_n),
\qquad
F_k^\star(t):=\Pbb(Y_k\le t).
\]
Define the corresponding $(1-\alpha)$ quantiles as
\[
q_{k,n}:=F_{k,n}^{-1}(1-\alpha),\qquad
q_k^\star:=F_k^{\star-1}(1-\alpha).
\]
Assume there exists an interval $\mathcal N$ containing both quantiles such that $F_k^\star$ admits a density $f_k^\star$ satisfying
\[
\inf_{t\in\mathcal N}f_k^\star(t)\ge f_{\min}>0,
\]
and assume $F_k^\star$ is globally Lipschitz with constant $L$. Then, for every training realization for which these conditions hold,
\begin{align}
|q_{k,n}-q_k^\star|
\le \frac{2\sqrt{2L}}{f_{\min}}
\sqrt{\Ebb_{\mathrm{cal}}\!\left[|X_{k,n}-Y_k|\mid\mathcal T_n\right]}.
\end{align}
\end{theorem}

\begin{proofe}
Condition on $\mathcal T_n$ throughout and suppress this conditioning in the notation. Let $p:=1-\alpha$ and $\Delta_F:=\sup_t|F_{k,n}(t)-F_k^\star(t)|$. Because $F_k^\star$ is continuous and has density at least $f_{\min}$ between the two quantiles, $F_k^\star(q_k^\star)=p$. If $q_{k,n}\ge q_k^\star$, then $F_{k,n}(t)<p$ for every $t<q_{k,n}$; taking $t\uparrow q_{k,n}$ and using continuity of $F_k^\star$ gives $F_k^\star(q_{k,n})\le p+\Delta_F$. Hence $f_{\min}(q_{k,n}-q_k^\star)\le\Delta_F$. If $q_{k,n}<q_k^\star$, then $F_{k,n}(q_{k,n})\ge p$, so $F_k^\star(q_{k,n})\ge p-\Delta_F$ and $f_{\min}(q_k^\star-q_{k,n})\le\Delta_F$. Thus,
\[
|q_{k,n}-q_k^\star|\le \Delta_F/f_{\min}.
\]
For any $\varepsilon>0$ and $t$,
\begin{align}
&\big|\mathds{1}\{X_{k,n}\le t\}-\mathds{1}\{Y_k\le t\}\big| \notag\\
&\le \mathds{1}\{|X_{k,n}-Y_k|\ge\varepsilon\}
+\mathds{1}\{t-\varepsilon<Y_k\le t+\varepsilon\}.
\end{align}
Taking conditional expectations, applying Markov's inequality, and using the global Lipschitz property of $F_k^\star$ gives
\[
|F_{k,n}(t)-F_k^\star(t)|
\le \frac{\Ebb_{\mathrm{cal}}[|X_{k,n}-Y_k|\mid\mathcal T_n]}{\varepsilon}+2L\varepsilon.
\]
If $\Ebb_{\mathrm{cal}}[|X_{k,n}-Y_k|\mid\mathcal T_n]=0$, then $X_{k,n}=Y_k$ almost surely and the claim is immediate. Otherwise, optimizing at
\[
\varepsilon=
\sqrt{\frac{\Ebb_{\mathrm{cal}}[|X_{k,n}-Y_k|\mid\mathcal T_n]}{2L}}
\]
yields the stated bound.$\hfill\IEEEQED$
\end{proofe}

\begin{remark}[Interpreting the constant]\label{rem:quantile-constant}
The factor $1/f_{\min}$ converts a CDF perturbation into a quantile perturbation, while $\sqrt L$ arises from the smoothing step that converts the conditional mean residual discrepancy into a uniform CDF bound. In particular, if $L\le1$,
\[
|q_{k,n}-q_k^\star|
\le \frac{2\sqrt2}{f_{\min}}
\sqrt{\Ebb_{\mathrm{cal}}[|X_{k,n}-Y_k|\mid\mathcal T_n]}.
\]
\end{remark}

\begin{corollary}[Quantile deviation via parameter error]\label{cor:quantile-ps}
Let
\[
L_{x,k}:=\sqrt{\Ebb\|\xbf_k^{\mathrm{cal}}\|_2^2},
\qquad
\widetilde{\wbf}_{k,n}:=\wbf_{k,n}-\wbf^\star.
\]
Assume the conditions of Theorem~\ref{thm:quantile-stability} hold almost surely over the training randomness with common constants $f_{\min}$ and $L$. Then, conditional on $\mathcal T_n$,
\begin{align}
|q_{k,n}-q_k^\star|
&\le \frac{2\sqrt{2L}}{f_{\min}}L_{x,k}^{1/2}
\|\widetilde{\wbf}_{k,n}\|_2^{1/2},\label{eq:quant-cond}
\end{align}
and consequently
\begin{align}
\Ebb_{\mathrm{tr}}|q_{k,n}-q_k^\star|
&\le \frac{2\sqrt{2L}}{f_{\min}}L_{x,k}^{1/2}
\Big(\Ebb_{\mathrm{tr}}\|\widetilde{\wbf}_{k,n}\|_2^2\Big)^{1/4}.
\label{eq:quant-msd}
\end{align}
\end{corollary}

\begin{proofe}
Conditional on $\mathcal T_n$, Lemma~\ref{lem:residual-lip} gives
\begin{align}
\Ebb_{\mathrm{cal}}[|X_{k,n}-Y_k|\mid\mathcal T_n]
&\le \Ebb\|\xbf_k^{\mathrm{cal}}\|_2\,
\|\widetilde{\wbf}_{k,n}\|_2 \\ 
&\le L_{x,k}\|\widetilde{\wbf}_{k,n}\|_2.
\end{align}
Substitution into Theorem~\ref{thm:quantile-stability} proves~\eqref{eq:quant-cond}. Taking expectation over training randomness and using
\[
\Ebb_{\mathrm{tr}}\|\widetilde{\wbf}_{k,n}\|_2^{1/2}
\le
\Big(\Ebb_{\mathrm{tr}}\|\widetilde{\wbf}_{k,n}\|_2^2\Big)^{1/4}
\]
proves~\eqref{eq:quant-msd}.$\hfill\IEEEQED$
\end{proofe}

\begin{corollary}[Quantile deviation via steady-state coefficient MSD]\label{cor:quantile-exact}
Under the conditions of Corollary~\ref{cor:quantile-ps}, if $\mathcal D_k$ in~\eqref{eq:steady-state-msd} exists, then
\begin{align}
\limsup_{n\to\infty}\Ebb_{\mathrm{tr}}|q_{k,n}-q_k^\star|
\le \frac{2\sqrt{2L}}{f_{\min}}L_{x,k}^{1/2}\mathcal D_k^{1/4}.
\end{align}
Thus, any training mechanism that yields a smaller steady-state coefficient MSD $\mathcal D_k$ tightens this expected quantile-deviation bound.
\end{corollary}

\begin{remark}[Coverage vs.\ efficiency]
The bounds above control deviations from a benign-reference quantile; they do not assert that a larger training error must move the realized quantile in a particular direction. In the studied attacks, degraded predictors empirically produce larger residuals and wider intervals, while configurations with lower training error generally yield tighter intervals. Conformal validity remains governed by the calibration construction under its exchangeability conditions.
\end{remark}

\begin{proposition}[Width perturbation bound]\label{prop:width}
Let $\Ccal(\xbf)=[\wbf^\intercal\xbf-q,\,\wbf^\intercal\xbf+q]$ have width $\omega=2q$, with $\omega_k^\star:=2q_k^\star$ and $\omega_{k,n}:=2q_{k,n}$. Define
\[
K_{\omega,k}:=\frac{4\sqrt{2L}}{f_{\min}}L_{x,k}^{1/2}.
\]
Under the conditions of Corollary~\ref{cor:quantile-ps}, conditional on $\mathcal T_n$,
\begin{align}
|\omega_{k,n}-\omega_k^\star|
&\le K_{\omega,k}\|\widetilde{\wbf}_{k,n}\|_2^{1/2},
\end{align}
and
\begin{align}
\Ebb_{\mathrm{tr}}|\omega_{k,n}-\omega_k^\star|
&\le K_{\omega,k}
\Big(\Ebb_{\mathrm{tr}}\|\widetilde{\wbf}_{k,n}\|_2^2\Big)^{1/4}.
\end{align}
\end{proposition}

\begin{proofe}
Since $\omega_{k,n}=2q_{k,n}$ and $\omega_k^\star=2q_k^\star$, both inequalities follow directly from Corollary~\ref{cor:quantile-ps}.$\hfill\IEEEQED$
\end{proofe}

\begin{theorem}[Per-iteration Byzantine attenuation and implications for width deviation]\label{theo:width-scaling-total}
Consider the training-stage attack model in Section~\ref{sec:training_attack}. Under A2--A3,
\begin{align}\label{eq:per-iteration-attenuation}
\Ebb\!\left[\|\beta_k\tau_{k,n}\Sbf_{k,n}\boldsymbol\delta_{k,n}\|_2^2\right]
=\beta_k p_a M\sigma_B^2.
\end{align}
For a Byzantine client, conditional on an attack occurring,
\begin{align}\label{eq:md_ratio}
\Ebb\!\left[\|\Sbf_{k,n}\boldsymbol\delta_{k,n}\|_2^2\mid\tau_{k,n}=1\right]=M\sigma_B^2,
\end{align}
whereas full sharing gives $D\sigma_B^2$. Hence the \emph{expected per-iteration} energy of this Byzantine client's perturbation contribution is reduced by the exact factor $M/D$ relative to full sharing.

Let $\mathcal D_k(M)$ denote the steady-state coefficient MSD under sharing level $M$, and define
\[
\Phi_k(M):=\limsup_{n\to\infty}\Ebb_{\mathrm{tr}}|\omega_{k,n}-\omega_k^\star|.
\]
Then Proposition~\ref{prop:width} gives
\begin{align}\label{eq:PhiM-bound}
\Phi_k(M)\le K_{\omega,k}\,\mathcal D_k(M)^{1/4}.
\end{align}
The exact $M/D$ attenuation in~\eqref{eq:per-iteration-attenuation} does not imply that $\mathcal D_k(M)$ scales monotonically with $M$, because both adversarial perturbations and optimization progress propagate through $M$-dependent learning dynamics. When the attenuation benefit yields a smaller $\mathcal D_k(M)$,~\eqref{eq:PhiM-bound} correspondingly tightens.
\end{theorem}

\begin{proofe}
Conditioning on $\Sbf_{k,n}$ and using its independence from $\boldsymbol\delta_{k,n}$,
\[
\Ebb\!\left[\|\Sbf_{k,n}\boldsymbol\delta_{k,n}\|_2^2\mid\Sbf_{k,n}\right]
=\sum_{d=1}^D[\Sbf_{k,n}]_{dd}^2\sigma_B^2=M\sigma_B^2.
\]
Taking expectations and using the independence of $\tau_{k,n}$ from $\Sbf_{k,n}$ and $\boldsymbol\delta_{k,n}$, together with $\beta_k^2=\beta_k$ and $\Ebb[\tau_{k,n}^2]=p_a$, gives~\eqref{eq:per-iteration-attenuation}. Equation~\eqref{eq:md_ratio} and the $M/D$ comparison follow immediately. The width bound~\eqref{eq:PhiM-bound} follows from Proposition~\ref{prop:width} and the definition of $\mathcal D_k(M)$.$\hfill\IEEEQED$
\end{proofe}

\subsection{Training Error and Calibration Robustness} \label{sec:calibration}

We now relate benign training error to calibration-stage Byzantine detection. For a fixed adversarial separation profile, a smaller benign predictor error makes the benign characterization vectors more tightly concentrated around their reference distribution and can therefore enlarge the guaranteed separation margin.

\begin{remark}[Common-reference assumption and non-IID extension]\label{rem:non-iid}
For the calibration-stage filtering analysis below, we impose the additional assumption that, under the reference predictor $\wbf^\star$, benign clients share a common raw-residual distribution $F^\star$. Thus, in the absence of predictor error, their population characterization vectors coincide with the common reference $\pbf^\star$; the benign population drift bounded in Lemma~\ref{lem:hist-lip} is attributed to predictor error. This common-reference assumption is stronger than A1--A3 and may be violated by intrinsic client heterogeneity, for example by client-dependent observation-noise laws. Our non-IID experiments in Section~\ref{sec:simulations} therefore assess the robustness of the filtering procedure empirically when this analytical assumption is relaxed.
\end{remark}

For the remainder of this subsection, we condition on the trained benign models and on a fixed normalization range and fixed histogram bin edges. Accordingly, the benign parameter errors are treated as fixed, and all probabilities below are with respect to the calibration samples and, where applicable, randomized Byzantine calibration submissions. Let $F^\star$ denote the benign raw-residual CDF under $\wbf^\star$. Benign client $k$ calibrates with $\widehat{\wbf}_k$ and parameter error $\ebf_k:=\widehat{\wbf}_k-\wbf^\star$; define
\[
e_{\max}:=\max_{k\in\Bcal}\|\ebf_k\|_2.
\]
For each benign client, let $L_{x,k}:=\sqrt{\Ebb\|\xbf_k\|_2^2}$ under its calibration distribution and define the uniform second-moment bound
\[
L_{x,\max}:=\max_{k\in\Bcal}L_{x,k}.
\]
Let $0=a_0<\cdots<a_H=1$ be the fixed normalized-score bin edges from Section~\ref{sec:calib-mitig}. Their equivalent raw-residual thresholds are $b_h:=R_{\max}a_h$ for $h=0,\ldots,H-1$ and $b_H:=\infty$. With the convention $F^\star(\infty)=1$, define the benign reference histogram by
\[
[\pbf^\star]_h:=F^\star(b_h)-F^\star(b_{h-1}),\qquad h=1,\ldots,H.
\]
This is equivalent to binning the clipped normalized scores using $I_1, \ldots, I_H$. Let $\qbf\in\mathbb R^H$ denote a fixed attack-dependent adversarial reference vector. Following Rob-FCP~\cite{kang2024certifiably}, define $d_{k,k'}:=\|\vbf_k-\vbf_{k'}\|_2$ and 
\[m_k:=\sum_{k'\in\mathrm{Far}(k,K_b-1)}d_{k,k'},\]
where $K_b:=K-|\Scal_B|$ is the number of benign clients. The server retains the $K_b$ clients with the smallest scores.

\begin{lemma}[Histogram concentration for benign clients]\label{lem:hist-concentration}
Let benign client $k$ have $N_k$ IID calibration residuals binned into the fixed raw-residual intervals $[b_{h-1},b_h)$ for $h<H$ and $[b_{H-1},\infty)$ for $h=H$, forming $\vbf_k\in\Delta^H$ with population vector $\pbf_k=\Ebb[\vbf_k]$. Then, for any $\delta\in(0,1)$, with probability at least $1-\delta$,
\[
\|\vbf_k-\pbf_k\|_2\le C_H\sqrt{\log(2H/\delta)/N_k},\qquad C_H=\sqrt{H/2}.
\]
\end{lemma}

\begin{proofe}
Let $\zbf_j\in\{e_1,\ldots,e_H\}$ denote the one-hot bin indicator of residual $j$. Each component of $\vbf_k=N_k^{-1}\sum_j\zbf_j$ is an average of independent Bernoulli variables. Hoeffding's inequality~\cite[Theorem 2.2.6]{vershynin2020high} and a union bound over $H$ bins give
\[
\Pbb\!\left(\|\vbf_k-\pbf_k\|_\infty>\varepsilon\right)\le2H\exp(-2N_k\varepsilon^2).
\]
Setting 
\[\varepsilon=\sqrt{\log(2H/\delta)/(2N_k)}\]
and then using $\|\vbf_k-\pbf_k\|_2\le\sqrt{H}\,\|\vbf_k-\pbf_k\|_\infty$ proves the claim.$\hfill\IEEEQED$
\end{proofe}

\begin{lemma}[Training error causes benign histogram drift]\label{lem:hist-lip}
Let $r(\wbf):=|y-\xbf^\intercal\wbf|$. For the fixed thresholds 
\[0=b_0<\cdots<b_{H-1}<b_H=\infty,\]
let $\pbf^\star$ and $\pbf_k$ denote the corresponding population histograms under $\wbf^\star$ and $\wbf^\star+\ebf_k$, respectively. Assume, uniformly over benign clients, that the conditional residual distribution of $r(\wbf^\star)$ given $\xbf$ admits a density $f_k^\star(u\mid\xbf)$ satisfying $\sup_{k,\xbf,u}f_k^\star(u\mid\xbf)\le f_{\max}$. Then
\begin{align}
\|\pbf_k-\pbf^\star\|_2\le4\sqrt H\,f_{\max}L_{x,k}\|\ebf_k\|_2.
\end{align}
Equivalently, 
\[\|\pbf_k-\pbf^\star\|_2\le C_{\mathrm{drift}}L_{x,k}\|\ebf_k\|_2\]
with $C_{\mathrm{drift}}:=4\sqrt H f_{\max}$.
\end{lemma}

\begin{proofe}
Let $r^\star=r(\wbf^\star)$ and $r_k=r(\wbf^\star+\ebf_k)$. By Lemma~\ref{lem:residual-lip}, 
\[|r_k-r^\star|\le\|\xbf\|_2\|\ebf_k\|_2=:\Delta(\xbf).\]
Conditioning on $\xbf$ and using the density bound gives, for finite $t$,
\[
|F_k(t\mid\xbf)-F_k^\star(t\mid\xbf)|\le2f_{\max}\Delta(\xbf).
\]
Averaging over $\xbf$, using the common-reference assumption $F_k^\star=F^\star$, and applying Cauchy--Schwarz yields
\[
|F_k(t)-F^\star(t)|\le2f_{\max}L_{x,k}\|\ebf_k\|_2.
\]
Define $g_h:=F_k(b_h)-F^\star(b_h)$ for $h=0,\ldots,H-1$ and $g_H:=0$ since $b_H=\infty$. Each bin-mass difference is $g_h-g_{h-1}$, so
\[
\|\pbf_k-\pbf^\star\|_2\le2\sqrt H\max_h|g_h|\le4\sqrt H f_{\max}L_{x,k}\|\ebf_k\|_2.
\]
This proves the claim.$\hfill\IEEEQED$
\end{proofe}

\begin{proposition}[Calibration separation under training error]\label{prop:margin}
Let $\Delta:=\|\qbf-\pbf^\star\|_2$, $N_{\min}:=\min_{k\in\Bcal}N_k$, and let $\mathcal E_a:=\{\max_{j\in\Scal_B}\|\vbf_j-\qbf\|_2\le r_a\}$. With probability at least $1-\delta$ over benign calibration sampling, on $\mathcal E_a$,
\begin{align}
\min_{j\in\Scal_B}\min_{k\in\Bcal}\|\vbf_j-\vbf_k\|_2\ge\Delta-r_a-r_b,
\end{align}
where
\[
r_b:=C_H\sqrt{\log(2HK_b/\delta)/N_{\min}}+C_{\mathrm{drift}}L_{x,\max} e_{\max}.
\]
For a fixed adversarial profile $(\Delta,r_a)$, any training method that yields a smaller $e_{\max}$ shrinks $r_b$ and increases the guaranteed separation margin $\Delta-r_a-r_b$.
\end{proposition}

\begin{proofe}[sketch]
For Byzantine $j$ and benign $k$, the reverse triangle inequality gives
\[
\|\vbf_j-\vbf_k\|_2\ge\Delta-\|\vbf_j-\qbf\|_2-\|\vbf_k-\pbf^\star\|_2.
\]
Lemma~\ref{lem:hist-concentration} with a union bound over $K_b$ benign clients and Lemma~\ref{lem:hist-lip}, together with $L_{x,k}\le L_{x,\max}$ and $\|\ebf_k\|_2\le e_{\max}$, give $\max_{k\in\Bcal}\|\vbf_k-\pbf^\star\|_2\le r_b$ with probability at least $1-\delta$. On $\mathcal E_a$, the remaining term is bounded by $r_a$, proving the claim.$\hfill\IEEEQED$
\end{proofe}

\begin{theorem}[Distance-based filtering guarantee]\label{thm:detection}
Consider the filtering rule in Section~\ref{sec:calib-mitig}, where the $B:=|\Scal_B|$ clients with the largest maliciousness scores are removed. Let $K_b:=K-B$, $N_{\min}:=\min_{k\in\Bcal}N_k$, $\Delta:=\|\qbf-\pbf^\star\|_2$, and
\begin{align}
r_b:=C_H\sqrt{\frac{\log(2HK_b/\delta)}{N_{\min}}}+C_{\mathrm{drift}}L_{x,\max} e_{\max}.\label{eq:rb-def}
\end{align}
Assume $\max_{j\in\Scal_B}\|\vbf_j-\qbf\|_2\le r_a$ with probability at least $1-\delta_a$ and $B<K_b-1$. Define
\[
\gamma:=\Delta-r_a-r_b,\qquad
\Gamma:=\Delta+r_a+r_b.
\]
If
\begin{align}
(K_b-1)\gamma>B\Gamma+(K_b-1-B)2r_b,\label{eq:score-sep}
\end{align}
then, conditional on the trained models and fixed binning specified above, the filtering is exact with probability at least $1-(\delta+\delta_a)$; hence $\Pbb(\text{misfiltering})\le\delta+\delta_a$. For fixed $(\Delta,r_a)$, a smaller $e_{\max}$ shrinks $r_b$ and makes~\eqref{eq:score-sep} easier to satisfy.
\end{theorem}

\begin{proofe}
See Appendix~\ref{app:detection-proof}.
\end{proofe}

\begin{remark}[Post-filtering calibration]\label{rem:postfilter}
On the exact-filtering event of Theorem~\ref{thm:detection}, Byzantine calibration submissions do not enter the conformal-quantile estimator. The remaining deviation from the ideal benign quantile is governed by benign predictor error, finite-sample calibration variability, and any approximation error introduced by the underlying FCP/sketching mechanism. On the complementary misfiltering event, no Byzantine-free quantile guarantee is claimed. When partial sharing yields a smaller benign training error in the studied attack-dominated regime, it correspondingly reduces $r_b$ and makes the exact-filtering condition easier to satisfy.
\end{remark}

\section{Simulation Results} \label{sec:simulations}

\subsection{Synthetic-Data Experiments}\label{sec:synth}

\emph{Experimental setup.} We consider a federated network of $K = 100$ clients. At each training iteration, the server uniformly samples $|\Scal_n| = 10$ clients to participate. Each client $k$ observes a non-IID stream $(\xbf_{k,n}, y_{k,n})$ generated by the linear model~\eqref{eq1}, with the ground-truth parameter vector normalized as $\|\wbf^\star\|_2=1$. 
We set the model dimension to $D = 50$ and apply partial sharing with $M = 15$ parameters per iteration (sharing ratio $M/D = 0.3$), which corresponds to a $70\%$ reduction in the number of exchanged parameters. The entries of $\xbf_{k,n}$ are drawn from Gaussian distributions $\Ncal (m_k, \varsigma_k^2)$ with client-specific means $m_k \sim \Ucal(-0.1,0.1)$ and variances $\varsigma_k^2 \sim \Ucal(0.2,1.2)$, while the observation noise is a zero-mean Gaussian with variance $\sigma_{\nu_k}^2 \sim \Ucal(0.005,0.025)$. 
The stepsize $\mu$ is set to $2.5 \times 10^{-2}$.
All results are averaged over $100$ independent Monte Carlo trials. 

\emph{Byzantine model.} We consider $|\Scal_B| = 20$ Byzantine clients (20\% of the network). During training, these clients inject additive Gaussian perturbations into their transmitted model updates with probability $p_a = 0.2$, using variance $\sigma_B^2 = 0.1$ (cf.\ Section~\ref{sec:training_attack}). 
During calibration, the same Byzantine clients submit adversarial nonconformity scores according to three attack scenarios (Fig.~\ref{fig:DiffAttacks}): (i) \emph{efficiency attack} (deflating scores to shrink intervals by sending all-zero scores), (ii) \emph{coverage attack} (inflating scores to widen intervals by sending scores $10\times$ the benign mean), and (iii) \emph{random attack} (adding zero-mean Gaussian noise with variance $\sigma_C^2 = 0.5$ to original scores while keeping them nonnegative). 
We set the size of the characterization vectors to $H = 100$ to balance histogram resolution and communication cost, and target $90\%$ coverage, i.e., $\alpha = 0.1$ in \eqref{eq:quantile}.

\begin{figure}[t!]
    \centering
    \includegraphics[width=0.475\textwidth]{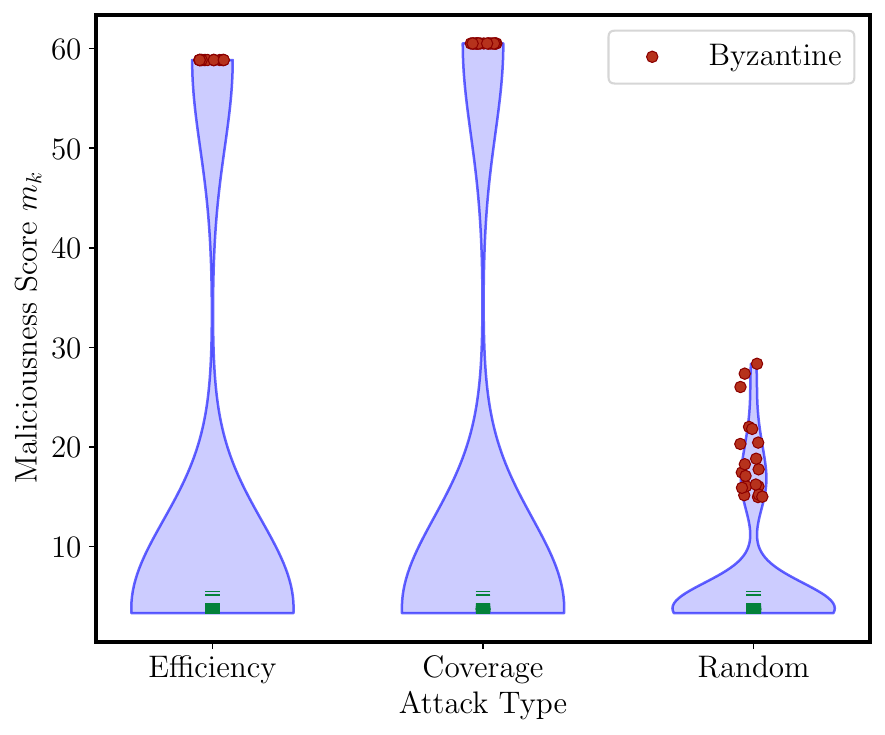}
    \caption{Distribution of maliciousness scores $m_k$ (cf.\ \eqref{eq:mal_score}) under different calibration-phase Byzantine attacks. Byzantine clients (\textcolor{BrickRed}{red}) attain markedly larger scores than benign clients, enabling reliable outlier filtering.}
    \label{fig:ViolinPlot}
\end{figure}

\emph{Maliciousness-score separation.} In Fig.~\ref{fig:ViolinPlot}, we show the maliciousness scores of all clients under the three calibration attacks. In each case, Byzantine clients concentrate on substantially larger scores than benign clients, making them clear outliers and enabling reliable filtering prior to quantile estimation. 
Note that random attacks present a particularly formidable challenge in the context of mitigation strategies. As the variance of the added Gaussian noise diminishes, specifically when $\sigma_C^2 \leq 0.2$, the difficulty associated with effectively countering these attacks increases. This phenomenon can be attributed to the increasingly covert nature of the random attacks, making their detection and subsequent mitigation notably more complex. 
Consequently, addressing the subtlety of these reduced-variance random attacks may require more sophisticated approaches in future work to strengthen robustness.

We present the marginal coverage and average prediction interval width for FCP, Rob-FCP, and PRISM-FCP under the same Byzantine attack scenarios as in Table~\ref{tab:synthetic}. We train for $1{,}000$ iterations, then perform conformal calibration using $1{,}000$ samples and evaluate on $1{,}000$ test samples per client. 
We illustrate how the prediction interval wraps around the target value in~\cite[Fig.~2]{lari2026partial}.

In Table~\ref{tab:synthetic}, we compare FCP, Rob-FCP, and PRISM-FCP under Byzantine attacks that affect both the training and calibration phases. Rob-FCP uses full model sharing ($M/D = 1$) with Byzantine filtering, while FCP uses full sharing without filtering. Under the coverage attacks, FCP exhibits inflated coverage ($100.0\%$) together with interval widths that are $4.4\times$ larger than PRISM-FCP, while both PRISM-FCP and Rob-FCP remain close to the nominal $90\%$ target. Under efficiency and random attacks, the three methods achieve similar coverage, but PRISM-FCP consistently produces tighter intervals, reflecting its lower training error. 
Overall, these results show that calibration-phase filtering is necessary for reliable uncertainty quantification under adversarial manipulation.

\begin{table}[t!]
\centering
\caption{Marginal coverage and average interval width under different Byzantine attacks ($K=100$ clients, $|\Scal_B|=20$ Byzantine, $D=50$, $\alpha=0.1$).}
\label{tab:synthetic}
\begin{tabular}{@{}llcc@{}}
\toprule
\textbf{Attack} & \textbf{Algorithm ($M/D$)} & \textbf{Cov. (\%)} & \textbf{Width} \\
\midrule
\multirow{3}{*}{Efficiency} 
    & PRISM-FCP (0.3) & $90.0 \pm 0.1$ & $1.76 \pm 0.09$ \\
    & Rob-FCP (1.0)   & $90.1 \pm 0.1$ & $2.04 \pm 0.12$ \\
    & FCP (1.0)       & $87.5 \pm 0.2$ & $1.88 \pm 0.11$ \\
\midrule
\multirow{3}{*}{Coverage} 
    & PRISM-FCP (0.3) & $90.0 \pm 0.1$ & $1.76 \pm 0.09$ \\
    & Rob-FCP (1.0)   & $90.1 \pm 0.1$ & $2.04 \pm 0.12$ \\
    & FCP (1.0)       & $100.0 \pm 0.0$& $7.77 \pm 0.54$ \\
\midrule
\multirow{3}{*}{Random} 
    & PRISM-FCP (0.3) & $90.0 \pm 0.1$ & $1.76 \pm 0.09$ \\
    & Rob-FCP (1.0)   & $90.0 \pm 0.1$ & $2.04 \pm 0.12$ \\
    & FCP (1.0)       & $91.7 \pm 0.6$ & $2.16 \pm 0.11$ \\
\bottomrule
\end{tabular}
\end{table}

PRISM-FCP further improves over Rob-FCP by substantially tightening the prediction intervals. Although both methods filter Byzantine clients during calibration, PRISM-FCP yields intervals that are approximately $14\%$ narrower than those of Rob-FCP by additionally mitigating Byzantine influence during training via partial sharing. This is consistent with Theorem~\ref{theo:width-scaling-total}, which establishes an $M/D$ attenuation of the expected per-iteration energy of each Byzantine client's perturbation contribution. The observed training MSE also improves from $-9.8\dB$ for Rob-FCP to $-13.6\dB$ for PRISM-FCP ($3.8\dB$ improvement). 

\begin{figure}[t!]
    \centering
    \vspace{-.8mm}
    \includegraphics[width=0.487\textwidth]{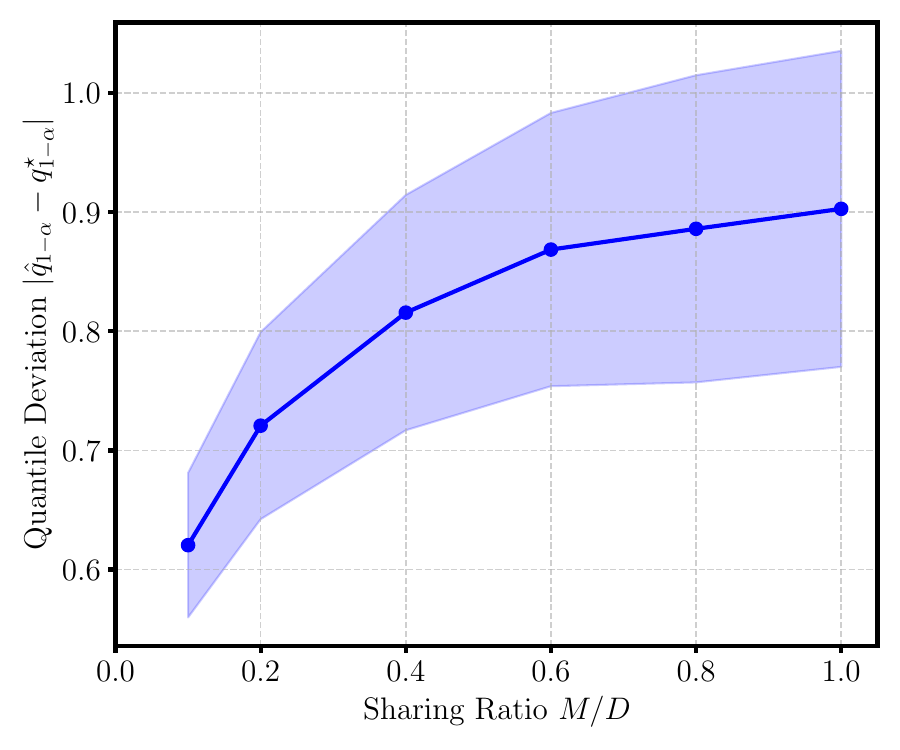}
    \caption{Quantile deviation $|\hat{q}_{1-\alpha} - q^\star_{1-\alpha}|$ of PRISM-FCP versus sharing ratio $M/D$ under the random attack.}
    \label{fig:quantile-deviation}
\end{figure}

Fig.~\ref{fig:quantile-deviation} is consistent with Corollary~\ref{cor:quantile-exact} and reports the empirical mean calibration-quantile deviation across sharing ratios. The deviation decreases monotonically as $M/D$ is reduced, and at $M/D = 0.1$ it is roughly $30\%$ smaller than under full sharing ($M/D = 1$). This trend is consistent with the attenuation mechanism in Lemma~\ref{lem:partial-sharing-attenuation} and with Corollary~\ref{cor:quantile-exact}: when partial sharing yields a smaller steady-state coefficient error, the derived bound on the conformal-quantile deviation tightens.

\emph{Robustness to unknown $|\Scal_B|$.} PRISM-FCP can be deployed either with a known Byzantine count $|\Scal_B|$, where score-based filtering excludes exactly $|\Scal_B|$ clients, or without this knowledge using a MAD-based outlier rule with a scale factor of $1.4826$ and a threshold of $2.5$ as its parameters. Across all three attack types, the known-$|\Scal_B|$ setting achieves perfect detection ($20/20$ true positives and \textit{no} false positives). 
When $|\Scal_B|$ is unknown, a practical alternative is to flag outliers using a robust MAD-based rule (Remark~\ref{rem:handling_unknown}). The MAD rule also attains $100\%$ recall, while incurring roughly \textit{one} false positive on average. Importantly, both settings preserve nominal coverage ($90\%$), indicating that PRISM-FCP remains robust even when $|\Scal_B|$ is not known a priori.

\subsection{UCI Superconductivity Experiments} \label{sec:real-world}

To demonstrate practical applicability, we evaluate PRISM-FCP on the UCI Superconductivity dataset~\cite{hamidieh2018superconductivity}, which contains $21{,}263$ samples with $D=81$ features describing material properties, with the target being the critical temperature, in Kelvin (K), at which superconductivity occurs.

\emph{Experimental Setup.}
We distribute the dataset across $K=100$ clients, including $|\Scal_B|=20$ Byzantine clients. Each client is assigned $3{,}000$ samples, split into $1{,}000$ training, $1{,}000$ calibration, and $1{,}000$ test samples.\footnote{To simulate non-IID data heterogeneity, we partition the target variable (critical temperature) into $10$ equal-frequency quantile bins and draw each client's mixing proportions over these bins from a $\mathrm{Dir}(0.5)$ distribution~\cite{hsu2019measuring}. Each client receives $3{,}000$ samples, with overlap permitted across clients.} This real-data construction is intended as a heterogeneity and robustness benchmark based on repeated random resampling from a finite dataset, rather than as a strictly disjoint federated partition of the UCI samples. At each training iteration, the server uniformly samples $|\Scal_n|=20$ clients to participate. Features are standardized to zero mean and unit variance, and targets are normalized similarly. As before, we compare: (i)~PRISM-FCP with partial sharing ($M<D$) and Byzantine filtering, (ii)~Rob-FCP with full sharing ($M=D$) and Byzantine filtering, and (iii)~FCP with full sharing and no filtering. We sweep sharing ratios $M/D \in \{0.25, 0.49, 0.74, 1.0\}$ (i.e., $M \in \{20, 40, 60, 81\}$) and average results over $100$ independent trials. The target coverage is $1-\alpha = 90\%$.

\begin{table}[t!]
\centering
\caption{Evaluation results on the UCI Superconductivity dataset with $D=81$.}
\label{tab:UCI}
\begin{tabular}{@{}llcc@{}}
\toprule
\textbf{Attack} & \textbf{Algorithm ($M/D$)} & \textbf{Cov. (\%)} & \textbf{Width (K)} \\
\midrule
\multirow{6}{*}{Efficiency} 
    & PRISM-FCP (0.25) & $90.0 \pm 0.1$ & $64.09 \pm 1.25$ \\
    & PRISM-FCP (0.49) & $90.0 \pm 0.1$ & $66.43 \pm 1.99$ \\
    & PRISM-FCP (0.74) & $90.0 \pm 0.2$ & $69.87 \pm 2.81$ \\
    & PRISM-FCP (1.0)  & $90.0 \pm 0.2$ & $74.33 \pm 4.37$ \\
    & Rob-FCP (1.0)    & $90.0 \pm 0.2$ & $74.33 \pm 4.37$ \\
    & FCP (1.0)        & $87.5 \pm 0.2$ & $69.36 \pm 4.10$ \\
\midrule
\multirow{6}{*}{Coverage} 
    & PRISM-FCP (0.25) & $90.0 \pm 0.1$ & $64.09 \pm 1.25$ \\
    & PRISM-FCP (0.49) & $90.0 \pm 0.1$ & $66.43 \pm 1.99$ \\
    & PRISM-FCP (0.74) & $90.0 \pm 0.2$ & $69.87 \pm 2.81$ \\
    & PRISM-FCP (1.0)  & $90.0 \pm 0.2$ & $74.33 \pm 4.37$ \\
    & Rob-FCP (1.0)    & $90.0 \pm 0.2$ & $74.33 \pm 4.37$ \\
    & FCP (1.0)        & $100.0 \pm 0.0$& $364.98 \pm 22.33$ \\
\midrule
\multirow{6}{*}{Random} 
    & PRISM-FCP (0.25) & $90.0 \pm 0.1$ & $64.33 \pm 1.28$ \\
    & PRISM-FCP (0.49) & $90.0 \pm 0.1$ & $66.66 \pm 2.04$ \\
    & PRISM-FCP (0.74) & $90.0 \pm 0.2$ & $70.09 \pm 2.86$ \\
    & PRISM-FCP (1.0)  & $90.0 \pm 0.2$ & $74.52 \pm 4.39$ \\
    & Rob-FCP (1.0)    & $90.0 \pm 0.2$ & $74.52 \pm 4.39$ \\
    & FCP (1.0)        & $91.7 \pm 0.3$ & $78.43 \pm 4.17$ \\
\bottomrule
\end{tabular}
\end{table}

In Table~\ref{tab:UCI}, we present marginal coverage and average interval width under the three calibration-phase attack scenarios. The results are consistent with the synthetic experiments and demonstrate the empirical performance of PRISM-FCP on real-world data:

\emph{Empirical coverage.} Across all three attacks and sharing ratios, PRISM-FCP remains close to the $90\%$ coverage target. In contrast, FCP without Byzantine filtering fails under adversarial calibration: it substantially inflates intervals under coverage attacks (reaching $100.0\%$ coverage with widths exceeding $ 5\times$) and undercovers under efficiency attacks (e.g., $87.5\%$ coverage).

\emph{Interval efficiency.} PRISM-FCP yields tighter intervals than Rob-FCP. For example, at $M/D=0.25$ ($M=20$), PRISM-FCP produces intervals that are about $13\%$ narrower than Rob-FCP across all three attacks. This empirical improvement is consistent with a reduction in training error when the attenuation of Byzantine influence from partial sharing outweighs the optimization slowdown caused by partial updates.

\emph{Communication-accuracy trade-off.} In this experiment, smaller sharing ratios generally yield tighter intervals, consistent with stronger Byzantine attenuation when it outweighs the slower optimization caused by partial updates. The benefit saturates around $M/D \approx 0.5$. In particular, reducing $M/D$ from $0.49$ to $0.25$ provides only marginal additional width reduction while slightly slowing convergence, suggesting $0.3 \leq M/D \leq 0.5$ as a practical operating range that balances efficiency and communication cost.

\emph{Attack-type invariance.} The relative advantage of PRISM-FCP over Rob-FCP is stable across efficiency, coverage, and random attacks, consistent with the interpretation that training-stage mitigation via partial sharing complements the calibration-phase filtering.

\emph{Sensitivity to the number of histogram bins $H$.} We investigate the sensitivity of PRISM-FCP to the number of histogram bins $H$ under the random calibration attack. Table~\ref{tab:H_sensitivity} reports the false-positive/false-negative counts, Byzantine detection rate, empirical coverage, and average interval width for $H \in \{25,50,100,200\}$ and sharing ratios $M/D \in \{0.25,0.49,0.74,1.00\}$. The empirical coverage remains stable across all tested values of $H$, staying within $[90.0\%,90.7\%]$. Increasing $H$ primarily improves detection accuracy: the average number of missed Byzantine clients decreases from about $5.44$ for $H=25$ to $0.47$ for $H=100$ and $0.06$ for $H=200$. Thus, $H=100$ already detects more than $97\%$ of Byzantine clients on average, while using only half the characterization dimension of $H=200$. Therefore, we use $H=100$ as a practical tradeoff between detection accuracy, histogram resolution, and communication overhead.

\begin{table}[t!]
\centering
\caption{Sensitivity of PRISM-FCP to the number of histogram bins \(H\) on the UCI Superconductivity dataset.}
\label{tab:H_sensitivity}
\setlength{\tabcolsep}{4pt}
\begin{tabular}{cccccc}
\toprule
\textbf{$H$} & \textbf{$M$} & \textbf{FP\,=\,FN} & \textbf{Rate (\%)} & \textbf{Coverage (\%)} & \textbf{Width (K)} \\
\midrule
\multirow{4}{*}{25}
 & 20 & $5.70 \pm 1.79$ & 71.5 & $90.7 \pm 0.31$ & $65.64 \pm 1.27$ \\
 & 40 & $5.25 \pm 1.92$ & 73.8 & $90.7 \pm 0.37$ & $67.82 \pm 2.01$ \\
 & 60 & $5.39 \pm 2.27$ & 73.1 & $90.7 \pm 0.35$ & $71.29 \pm 2.72$ \\
 & 81 & $5.43 \pm 2.72$ & 72.9 & $90.7 \pm 0.35$ & $75.25 \pm 4.00$ \\
\midrule
\multirow{4}{*}{50}
 & 20 & $2.30 \pm 1.37$ & 88.5 & $90.3 \pm 0.29$ & $65.09 \pm 1.32$ \\
 & 40 & $1.99 \pm 1.40$ & 90.1 & $90.3 \pm 0.32$ & $67.12 \pm 1.97$ \\
 & 60 & $1.98 \pm 1.41$ & 90.1 & $90.3 \pm 0.26$ & $70.51 \pm 2.64$ \\
 & 81 & $2.02 \pm 1.67$ & 89.9 & $90.3 \pm 0.28$ & $74.49 \pm 3.94$ \\
\midrule
\multirow{4}{*}{100}
 & 20 & $0.61 \pm 0.68$ & 96.9 & $90.1 \pm 0.24$ & $64.68 \pm 1.38$ \\
 & 40 & $0.43 \pm 0.59$ & 97.8 & $90.1 \pm 0.22$ & $66.65 \pm 1.95$ \\
 & 60 & $0.38 \pm 0.63$ & 98.1 & $90.1 \pm 0.20$ & $70.04 \pm 2.62$ \\
 & 81 & $0.45 \pm 0.79$ & 97.8 & $90.1 \pm 0.20$ & $74.06 \pm 3.99$ \\
\midrule
\multirow{4}{*}{200}
 & 20 & $0.11 \pm 0.34$ & 99.5 & $90.0 \pm 0.19$ & $64.47 \pm 1.32$ \\
 & 40 & $0.05 \pm 0.22$ & 99.7 & $90.0 \pm 0.17$ & $66.48 \pm 1.93$ \\
 & 60 & $0.02 \pm 0.14$ & 99.9 & $90.0 \pm 0.16$ & $69.89 \pm 2.62$ \\
 & 81 & $0.07 \pm 0.29$ & 99.7 & $90.0 \pm 0.14$ & $73.92 \pm 3.92$ \\
\bottomrule
\end{tabular}
\end{table}

\subsection{Incorporation of Robust Aggregation Methods}

We evaluate PRISM-FCP against five full-sharing training baselines, namely Rob-FCP with FedAvg~\cite{mcmahan2017communication}, Coordinate-Wise Median (CWMed)~\cite{yin2018byzantine}, Coordinate-Wise Trimmed Mean (CWTrMean)~\cite{yin2018byzantine}, Krum~\cite{blanchard2017machine}, and Multi-Krum~\cite{blanchard2017machine}. To ensure a fair comparison regarding calibration robustness, all baselines are combined with the same histogram-based calibration filtering. We also combine partial sharing ($M/D=0.2$) with CWMed aggregation. As reported in Table~\ref{tab:baseline_gaussian_results}, full-sharing CWMed achieves a steady-state MSE of $-23.48$~dB, while adding partial sharing improves the MSE to $-23.84$~dB. The interval width is slightly larger than full-sharing CWMed, but the communicated model-parameter payload is reduced by $80\%$ relative to full sharing. Thus, partial sharing should be viewed as complementary to robust aggregation rather than as a replacement for it.

\begin{table*}[t!]
\centering
\caption{Performance of PRISM-FCP versus robust aggregation baselines under Gaussian Byzantine attack.}
\label{tab:baseline_gaussian_results}
\begin{tabular}{@{}lcccccccc@{}}
\toprule
\multirow{2}{*}{\textbf{Algorithm}} & \multirow{2}{*}{\textbf{$\cfrac{M}{D}$}} & \multirow{2}{*}{\textbf{MSE ($\dB$)}} & \multicolumn{2}{c}{\textbf{Efficiency Attack}} & \multicolumn{2}{c}{\textbf{Coverage Attack}} & \multicolumn{2}{c}{\textbf{Random Attack}} \\
\cmidrule(lr){4-5} \cmidrule(lr){6-7} \cmidrule(l){8-9}
 & & & Cov. (\%) & Width & Cov. (\%) & Width & Cov. (\%) & Width \\
\midrule
PRISM-FCP & 0.2 & -13.59 & 90.0 & 1.6230 & 90.0 & 1.6230 & 90.0 & 1.6241 \\
PRISM-FCP & 0.4 & -12.38 & 90.0 & 1.8446 & 90.0 & 1.8446 & 90.0 & 1.8468 \\
PRISM-FCP & 0.6 & -11.46 & 90.0 & 1.9524 & 90.0 & 1.9524 & 90.1 & 1.9558 \\
PRISM-FCP & 0.8 & -10.67 & 90.0 & 2.0006 & 90.0 & 2.0006 & 90.1 & 2.0044 \\
Rob-FCP   & 1.0 & -9.81  & 90.0 & 2.0416 & 90.0 & 2.0416 & 90.1 & 2.0486 \\
PSO-CWMed & 0.2 & -23.84 & 90.0 & 0.4969 & 90.0 & 0.4969 & 90.0 & 0.4969 \\
CWMed     & 1.0 & -23.48 & 90.0 & 0.4404 & 90.0 & 0.4404 & 90.0 & 0.4404 \\
CWTrMean  & 1.0 & -22.99 & 90.0 & 0.4616 & 90.0 & 0.4616 & 90.0 & 0.4616 \\
Krum      & 1.0 & -18.24 & 90.0 & 0.7692 & 90.0 & 0.7692 & 90.0 & 0.7692 \\
Multi-Krum& 1.0 & -20.53 & 90.0 & 0.5833 & 90.0 & 0.5833 & 90.0 & 0.5833 \\
PSO-FCP   & 0.2 & -13.59 & 87.6 & 1.4991 & 100.0& 6.4661 & 92.4 & 1.7644 \\
FCP       & 1.0 & -9.81  & 87.5 & 1.8862 & 100.0& 7.7755 & 91.7 & 2.1648 \\
\bottomrule
\end{tabular}
\end{table*}

\subsection{Stealthy Byzantine Attacks}

To evaluate the robustness of the proposed framework beyond simple Gaussian perturbations, we additionally consider the stealthy, distribution-aware A Little Is Enough (ALIE) attack~\cite{baruch2019little}. Unlike Gaussian noise, ALIE constructs malicious updates that lie within the expected coordinate-wise variation of benign updates, making the attack difficult to detect using standard robust aggregation rules. 
For this experiment, the client-specific means are drawn from $m_k \sim \Ucal(-2,2)$.

As reported in Table~\ref{tab:alie_results}, ALIE degrades several robust aggregation baselines, particularly Krum and Multi-Krum. PRISM-FCP maintains near-nominal conformal coverage and competitive interval width while sharing only $20\%$ of the model coordinates per round. In addition, PRISM-FCP provides a favorable communication--robustness tradeoff, reducing the communicated model-parameter payload by $80\%$ relative to full-sharing baselines. 
These results suggest that partial model sharing can remain useful under this stealthy, distribution-aware attack, while we do not claim protection against fully adaptive Byzantine strategies.
In other words, when communication is not a limiting factor, Rob-FCP may be preferable under this particular attack, while in communication-constrained federated deployments, PRISM-FCP offers a substantially lower-communication alternative while retaining robust conformal performance.

\begin{table*}[t!]
\centering
\caption{Performance of PRISM-FCP versus various robust aggregation baselines under the ALIE attack.}
\label{tab:alie_results}
\begin{tabular}{@{}llccccccc@{}}
\toprule
\multirow{2}{*}{\textbf{Algorithm}} & \multirow{2}{*}{\textbf{$\cfrac{M}{D}$}} & \multirow{2}{*}{\textbf{MSE ($\dB$)}} & \multicolumn{2}{c}{\textbf{Efficiency Attack}} & \multicolumn{2}{c}{\textbf{Coverage Attack}} & \multicolumn{2}{c}{\textbf{Random Attack}} \\
\cmidrule(lr){4-5} \cmidrule(lr){6-7} \cmidrule(l){8-9}
 & & & Cov. (\%) & Width  & Cov. (\%) & Width  & Cov. (\%) & Width  \\
\midrule
PRISM-FCP & 0.2 & -18.84 & 89.9 & 0.5063 & 89.9 & 0.5063 & 89.9 & 0.5063 \\
Rob-FCP   & 1.0 & -19.97 & 89.9 & 0.4349 & 89.9 & 0.4344 & 92.9 & 0.4834 \\
CWMed     & 1.0 & -19.04 & 89.9 & 0.5640 & 89.8 & 0.5622 & 95.2 & 0.7158 \\
CWTrMean  & 1.0 & -19.04 & 89.9 & 0.5640 & 89.8 & 0.5622 & 95.2 & 0.7158 \\
Krum      & 1.0 & -17.14 & 89.4 & 0.7953 & 89.8 & 0.8028 & 95.5 & 1.0345 \\
Multi-Krum& 1.0 & -18.78 & 89.9 & 0.6011 & 89.8 & 0.5990 & 96.0 & 0.7990 \\
\bottomrule
\end{tabular}
\end{table*}

\subsection{Sign-Flipping Byzantine Attacks}

Using the synthetic-data setup of Section~\ref{sec:synth}, with the modifications specified below, we further evaluate PRISM-FCP under a sign-flipping model-poisoning attack, in which Byzantine clients reverse and amplify their local updates. Specifically, letting $\mathbf{g}$ denote the local stochastic gradient, a Byzantine client applies the update $\mathbf{w}+\varepsilon\mu\mathbf{g}$ instead of the benign descent update $\mathbf{w}-\mu\mathbf{g}$~\cite{blanchard2017machine}. The data-generation and optimization settings otherwise follow Section~\ref{sec:synth}, and all compared methods use the same histogram-based conformal-calibration pipeline and nominal $90\%$ coverage target.

Table~\ref{tab:signflip} reports the results for $\varepsilon=1.1$ under a challenging setting with $|\Scal_B|=40$ Byzantine clients (40\% of the network), twice the Byzantine fraction used in our main Gaussian-attack experiments, averaged over 100 Monte Carlo trials. Under this attack, the full-sharing Rob-FCP/FedAvg baseline \emph{diverges}, whereas PRISM-FCP with $M/D=0.2$ remains stable, achieving a steady-state MSE of $-17.69$~dB, 89.9\% empirical coverage for a 90\% target, and an average interval width of $0.525$. Its MSE is within $0.26$~dB of full-sharing CWMed and $0.56$~dB of Multi-Krum, while communicating only 20\% of the model coordinates per round. These results are consistent with the intuition that restricting the coordinates communicated in each round can limit the instantaneous influence of malicious updates, although we do not claim a general robustness guarantee against sign-flipping attacks.

\begin{table}[t!]
\centering
\caption{Performance under the sign-flipping attack ($\varepsilon = 1.1$, $|\Scal_B| = 40$, $K = 100$, 100 Monte Carlo trials).}
\label{tab:signflip}
\begin{tabular}{lcccc}
\toprule
\textbf{Algorithm} & \textbf{MSE (dB)} & \textbf{Cov. (\%)} & \textbf{Width} & \textbf{Quant. Dev.} \\
\midrule
PRISM-FCP & $-17.69$ & $89.9$ & $0.525$ & $0.061 \pm 0.028$ \\
CWMed & $-17.95$ & $90.0$ & $0.501$ & $0.048 \pm 0.017$ \\
CWTrMean & $-17.95$ & $90.0$ & $0.501$ & $0.048 \pm 0.017$ \\
Krum & $-14.68$ & $90.0$ & $0.739$ & $0.168 \pm 0.062$ \\
Multi-Krum & $-18.25$ & $90.0$ & $0.483$ & $0.039 \pm 0.018$ \\
Rob-FCP & \multicolumn{4}{c}{\emph{Diverged}} \\
\bottomrule
\end{tabular}
\end{table}

\subsection{Nonlinear Regression with Neural Networks}

To assess PRISM-FCP at a substantially larger nonlinear model and data scale, we evaluate it on the UCI YearPredictionMSD dataset~\cite{bertin2011year}. The dataset contains $515{,}345$ songs, with $463{,}715$ instances designated for training and $51{,}630$ for testing. Each instance comprises $90$ continuous audio-timbre features (12 timbre PCA means and 78 covariances), and the regression task is to predict the song's release year.

\emph{Architecture and Partitioning.} We deploy a multi-layer perceptron (MLP) with four affine layers of dimensions $90 \to 512 \to 256 \to 128 \to 1$ and ReLU activations, comprising $D = 210{,}945$ trainable parameters. Across $K = 10$ clients, data are partitioned under a non-IID Dirichlet distribution with concentration parameter $\alpha_{\mathrm{Dir}} = 0.5$ across $10$ target quantile bins. Each client receives disjoint, non-overlapping partitions containing $40{,}000$ training, $6{,}371$ calibration, and $5{,}163$ test samples, so that $515{,}340$ of the $515{,}345$ available instances are used. Input features are standardized via a scaler fitted strictly on the training data, and targets are standardized using training-set statistics.

\emph{Optimization and Mask Synchronization.} All $K = 10$ clients participate in every round. Clients train locally using AdamW (weight decay $10^{-4}$, batch size $B = 128$) for $E = 20$ local steps per round over $T = 35$ communication rounds. The learning rate follows a cosine annealing schedule from $\eta_0 = 10^{-3}$ to a positive lower floor $\eta_{\min} = 4 \times 10^{-4}$. At each round $t$, a coordinate mask $\mathbf{m}_t \in \{0, 1\}^D$ is generated via uniform sampling without replacement from a synchronized pseudo-random generator, so no per-round transmission of coordinate indices is required. Only the $M$ active parameters are exchanged in the uplink and downlink, yielding an exact reduction of $(1 - M/D) \times 100\%$ in the communicated model-parameter payload relative to full sharing. This synchronized-mask neural-network experiment is an empirical extension; the client-wise mask independence in A3 is specific to the linear theoretical analysis.

\emph{Byzantine Attacks and Inference Protocol.} We instantiate $|\mathcal{S}_B| = 2$ Byzantine clients ($20\%$ adversarial fraction). During training, adversaries perform attacks with probability $p_a = 0.2$, corrupting coordinates with zero-mean Gaussian noise of variance $\sigma_B^2 = 0.1$. We also evaluate ALIE with parameter $z = 1.0$ to assess performance under stealthy distribution-aware poisoning. Configurations compared across sharing ratios use matched Monte Carlo seeds and, where applicable, matched attack schedules and underlying random draws across the five trials. 
In the calibration phase, each client $k$ computes nonconformity scores $s_{k, i} = |y_{k, i}^{\mathrm{cal}} - \hat{f}_k(\mathbf{x}_{k, i}^{\mathrm{cal}})|$ on its local calibration set using its local model $\hat{f}_k$, and constructs both an $H=100$-bin characterization vector and a score summary for quantile estimation. Byzantine clients manipulate their calibration submissions according to the considered attack; for the Random calibration attack, the added Gaussian noise has variance $\sigma_C^2=1.0$. The server uses the characterization vectors to identify and filter Byzantine clients, estimates the global $(1-\alpha)$ conformal quantile from only the score summaries of the retained clients, and broadcasts the resulting $\hat q_{1-\alpha}$ to all clients. Each client then constructs its prediction intervals using its local model and evaluates them on its local test set.

\begin{table}[t!]
\centering
\caption{Performance evaluation on the UCI YearPredictionMSD dataset ($K=10$ clients, $|\mathcal{S}_B|=2$ Byzantine, MLP with four affine layers and $D = 210{,}945$ parameters, nominal coverage target $1-\alpha=90\%$).}
\label{tab:yearprediction_sweep}
\setlength{\tabcolsep}{2.5pt}
\begin{tabular}{cccccc}
\toprule
\textbf{Tr. Att.} & \textbf{Algorithm} & \textbf{$M/D$} & \textbf{Cov. (\%)} & \textbf{Width (yr)} & \textbf{RMSE (yr)}\\
\midrule
None & FedAvg & $1.0$ & $89.7 \pm 0.20$ & $26.10 \pm 3.10$ & $8.96 \pm 0.21$\\
\midrule
\multirow{5}{*}{Gaussian}
 & \multirow{4}{*}{PRISM-FCP} & $0.3$ & $89.8 \pm 0.20$ & $27.62 \pm 3.54$ & $9.56 \pm 0.25$\\
 &  & $0.5$ & $89.8 \pm 0.19$ & $27.80 \pm 3.54$ & $9.53 \pm 0.29$\\
 &  & $0.7$ & $89.7 \pm 0.18$ & $28.21 \pm 3.57$ & $9.49 \pm 0.33$\\
 &  & $0.9$ & $89.7 \pm 0.22$ & $28.69 \pm 3.82$ & $9.52 \pm 0.31$\\
 & Rob-FCP   & $1.0$ & $89.6 \pm 0.17$ & $28.85 \pm 4.05$ & $9.59 \pm 0.35$\\
\midrule
\multirow{2}{*}{ALIE}
 & PRISM-FCP & $0.5$ & $89.7 \pm 0.20$ & $26.84 \pm 3.46$ & $9.15 \pm 0.18$\\
 & Rob-FCP   & $1.0$ & $89.7 \pm 0.37$ & $26.65 \pm 3.26$ & $9.16 \pm 0.24$\\
\bottomrule
\end{tabular}
\end{table}

All metrics report the mean $\pm$ standard deviation across five independent Monte Carlo seeds ($\{0, 42, 123, 1234, 2026\}$). We evaluate steady-state test root mean-square error (RMSE) on the unnormalized scale in calendar years over the final $10$ rounds ($t \in [26, 35]$). 
The first row provides an attack-free FedAvg reference. For the Byzantine-attack configurations, Table~\ref{tab:yearprediction_sweep} reports results for the Random calibration attack ($\sigma_C^2=1.0$). The Efficiency and Coverage attacks yield identical post-filtering metrics because the server removes both Byzantine clients in every evaluated trial, yielding the same retained benign calibration set. Under Gaussian training-stage poisoning, reducing the sharing ratio modestly decreases the mean interval width while maintaining empirical coverage close to the nominal target. At $M/D=0.3$, PRISM-FCP achieves the smallest mean interval width among the tested Gaussian-attack configurations while reducing the communicated model-parameter payload by $70\%$ relative to full sharing. Under ALIE, PRISM-FCP with $M/D=0.5$ achieves coverage and interval width comparable to full-sharing Rob-FCP while reducing the communicated model-parameter payload by approximately $50\%$. Training remains empirically stable across the tested sharing ratios.

\section{Conclusion and Future Work} \label{sec:conclusion}

We proposed PRISM-FCP to mitigate Byzantine attacks in both the training and calibration phases of federated conformal prediction. Under the stated stochastic training-stage attack model, we showed that partial sharing attenuates the expected per-iteration energy of each Byzantine client's perturbation contribution by $M/D$ relative to full sharing. When this attenuation yields a smaller steady-state coefficient error, it tightens the derived bounds on residual-quantile and interval-width deviations, while histogram-based filtering mitigates adversarial calibration submissions. Experiments on synthetic and real-world datasets are consistent with the theoretical trends in the analyzed setting and demonstrate favorable empirical communication--performance tradeoffs, including on the substantially larger nonlinear YearPredictionMSD benchmark and under
Gaussian, ALIE, and sign-flipping training-stage attacks.
Several directions merit future investigation. Extending the theoretical analysis beyond linear regression to nonlinear models and larger neural networks is an important next step. 
Analyzing robustness against fully adaptive adversaries, particularly mask-aware attackers that exploit partial-sharing patterns, is also important. Potential defenses include private or otherwise adversary-unpredictable sharing patterns and their integration with robust aggregation.
Extending PRISM-FCP to classification would require classification-specific nonconformity scores and set-valued prediction rules, such as Adaptive Prediction Sets~\cite{romano2020classification}, and is left for future work.

\appendices

\section{Proof of Theorem~\ref{thm:detection}}
\label{app:detection-proof}
\begin{proofe}
Condition on the trained benign models and a fixed normalization/binning rule, as in Theorem~\ref{thm:detection}. For each benign client $k\in\Bcal$, Lemma~\ref{lem:hist-concentration} with failure probability $\delta/K_b$ gives
\[
\Pbb\!\left(\|\vbf_k-\pbf_k\|_2>
C_H\sqrt{\frac{\log(2HK_b/\delta)}{N_k}}\right)\le\frac{\delta}{K_b}.
\]
Since $N_k\ge N_{\min}$, a union bound over the $K_b$ benign clients, followed by Lemma~\ref{lem:hist-lip}, $L_{x,k}\le L_{x,\max}$, and $\|\ebf_k\|_2\le e_{\max}$, yields, with probability at least $1-\delta$,
\begin{align*}
\max_{k\in\Bcal}\|\vbf_k-\pbf^\star\|_2
&\le C_H\sqrt{\frac{\log(2HK_b/\delta)}{N_{\min}}}
+C_{\mathrm{drift}}L_{x,\max}e_{\max}\\
&=r_b.
\end{align*}
Denote this event by $\mathcal E_b$. By assumption, 
\[\mathcal E_a:=\{\max_{j\in\Scal_B}\|\vbf_j-\qbf\|_2\le r_a\}\]
has probability at least $1-\delta_a$. Hence, the joint event satisfies
\[
\mathcal E:=\mathcal E_b\cap\mathcal E_a,\qquad
\Pbb(\mathcal E)\ge1-(\delta+\delta_a).
\]

Condition now on $\mathcal E$. For benign $k,k'$, the triangle inequality gives 
\[\|\vbf_k-\vbf_{k'}\|_2\le\|\vbf_k-\pbf^\star\|_2+\|\vbf_{k'}-\pbf^\star\|_2\le2r_b.\]
For Byzantine $j$ and benign $k$, the reverse triangle inequality and triangle inequality give, respectively,
\begin{align}
\|\vbf_j-\vbf_k\|_2
&\ge \Delta-\|\vbf_j-\qbf\|_2
-\|\vbf_k-\pbf^\star\|_2 \notag\\
&\ge \Delta-r_a-r_b=\gamma, \notag\\
\|\vbf_j-\vbf_k\|_2
&\le \|\vbf_j-\qbf\|_2+\Delta
+\|\vbf_k-\pbf^\star\|_2 \notag\\
&\le \Delta+r_a+r_b=\Gamma.
\label{eq:app-cross-dist}
\end{align}

Fix a Byzantine client $j$. All $K_b$ benign clients are at a distance of at least $\gamma$ from $j$. Since $\mathrm{Far}(j,K_b-1)$ contains the $K_b-1$ largest distances from $j$, its sum is at least the sum over any $K_b-1$ benign clients, and hence
\[
m_j\ge(K_b-1)\gamma.
\]
For a benign client $k$, condition~\eqref{eq:score-sep} has a nonnegative right-hand side and thus implies
\[
\gamma>0\ \Longrightarrow\ \Delta>r_a+r_b\ \Longrightarrow\ \Gamma=\Delta+r_a+r_b>2r_b.
\] Among the $K_b-1$ farthest neighbors of $k$, at most $B$ can be Byzantine. Using~\eqref{eq:app-cross-dist} for Byzantine neighbors and the bound $2r_b$ for benign neighbors gives
\[
m_k\le B\Gamma+(K_b-1-B)2r_b.
\]
Combining the two score bounds with~\eqref{eq:score-sep} gives
\[
m_j>m_k,\qquad\forall\,j\in\Scal_B,\quad \forall\,k\in\Bcal.
\]
Thus the $B$ largest maliciousness scores belong exactly to the Byzantine clients, so filtering is exact on $\mathcal E$. Since misfiltering can occur only on $\mathcal E^c$, 
\[\Pbb(\text{misfiltering})\le\Pbb(\mathcal E^c)\le\delta+\delta_a.\tag*{\IEEEQED}\]
\end{proofe}

\bibliographystyle{IEEEtran}
\bibliography{refs}

\end{document}